\documentclass{article}
\usepackage{iclr2027_conference,times}
\setcitestyle{numbers,square,comma}
\usepackage{hyperref}
\usepackage{tabularx}
\usepackage{booktabs}
\usepackage{makecell}
\usepackage{array}
\usepackage{amsmath,amssymb,amsthm,mathtools}    
\usepackage{nicefrac}           
\usepackage{xcolor}    
\usepackage{graphicx} 
\usepackage{xurl}
\usepackage{subcaption}
\usepackage{wrapfig}
\makeatletter
\newcommand{\wrapfill}{%
  \par
  \ifx\parshape\WF@fudgeparshape
    \nobreak
    \ifnum\c@WF@wrappedlines>\@ne
      \advance\c@WF@wrappedlines\m@ne
      \vskip\c@WF@wrappedlines\baselineskip
      \global\c@WF@wrappedlines\z@
    \fi
    \allowbreak
    \WF@finale
  \fi
}
\makeatother

\title{Shapley-based Data Valuation for LLM Alignment via Sequential Preference Optimization}
\author{Mélissa Tamine \thanks{Corresponding author.} \\
Criteo AI lab, FairPlay joint team, France \\
CREST, ENSAE, Institut Polytechnique de Paris \\
\texttt{m.tamine@criteo.com} \\
\AND
Otmane Sakhi \& Benjamin Heymann \& Maxime Vono \\
Criteo AI Lab, FairPlay joint team, France \\
\texttt{\{o.sakhi,b.heymann,m.vono\}@criteo.com} \\
\AND
Patrick Loiseau \\
Inria, FairPlay joint team, France \\
\texttt{patrick.loiseau@inria.fr}
}

\iclrfinalcopy 
\definecolor{mydarkblue}{rgb}{0,0.08,0.45}
\hypersetup{
    colorlinks=true,
    linkcolor=mydarkblue,
    citecolor=mydarkblue,
    filecolor=mydarkblue,
    urlcolor=mydarkblue,
    pdfview=FitH
}

\newtheorem{lemma}{Lemma}[section]
\newtheorem{assumption}{Assumption}[section]
\newtheorem{proposition}{Proposition}[section]

\newcommand{\X}{\mathcal X}
\newcommand{\Y}{\mathcal Y}
\newcommand{\D}{\mathcal D}
\newcommand{\N}{\mathcal N}
\newcommand{\E}{\mathbb E}

\newcommand{\argmin}{\operatorname*{arg\,min}}
\newcommand{\argmax}{\operatorname*{arg\,max}}
\begin{document}

\maketitle
\lhead{Preprint.}

\begin{abstract}
Data valuation is a natural framework for understanding which data sources matter most when aligning a Large Language Model (LLM) from multiple sources. The standard game-theoretic approach treats each source, or equivalently each preference dataset, as a player in a cooperative game and assigns it a contribution score through the Shapley value. In practice, however, Shapley-based valuation is computationally prohibitive because it requires aligning a separate model for every possible coalition of sources, i.e., an exponential number of alignments. We address this challenge for Direct Alignment Algorithms (DAAs), including IPO, which learn through log-policy ratios with respect to a reference policy. We show that, when a model is aligned sequentially source by source, exact optimization makes each stage contribute additively to the log-probability of a full response, up to a prompt-dependent normalization constant. This allows the log-probability assigned by any coalition to a fixed response to be reconstructed from the base policy and the policies trained on each source individually. This reduces the alignment cost of Shapley-based valuation from exponential to linear, since only one model per source needs to be trained to evaluate coalition scores. We test whether this theoretical property remains approximately valid under finite training across several base models and real-world data sources. We finally compute the Shapley values of these sources under multiple reward models, showing how their estimated contributions vary across evaluation criteria.
\end{abstract}
\section{Introduction}
\label{sec:introduction}
Large language models (LLMs) are aligned using preference data drawn from multiple heterogeneous sources. These sources may differ in the prompts they cover or the procedures used to construct them, but they are ultimately combined to shape the aligned model's behavior. Preference-based alignment methods, including Reinforcement Learning from Human Feedback (RLHF) \citep{christiano2017,ziegler2019,jaques2017} and direct preference-optimization methods such as Direct Preference Optimization (DPO) \citep{rafailov2023} and Identity Preference Optimization (IPO) \citep{azar2024}, make such data a central component of the alignment pipeline. This raises a natural \emph{data valuation} question: \emph{how should we attribute the contribution of each preference-data source to the behavior of the aligned model?} 
Answering this question is relevant for data markets (\emph{what is a fair price for a dataset?}), incentive design (\emph{how should contributors be rewarded?}), and responsibility attribution (\emph{which sources are responsible for desirable or undesirable model behavior?}) \citep{agarwal2019, chen2016, koh2017, sim2020, ghorbani2019, jia2019a, jia2019b}

Cooperative game theory, and in particular the Shapley value \citep{shapley1953}, provides a formalism for such data valuation problems by treating each data source as a player, assigning a utility to every coalition of players, and splitting the resulting utility among sources. However, directly instantiating this paradigm for LLM alignment is computationally prohibitive. The utility must be evaluated for every coalition of sources, and the number of coalitions grows exponentially with the number of sources \citep{shapley1953, ghorbani2019}. Even in classical supervised learning, this already leads to an impractical number of model retrainings. For LLM alignment, each coalition evaluation may require a full preference-optimization run, making the standard Shapley procedure infeasible at the scale of modern alignment pipelines. Emulation or distillation techniques can reduce the cost of each run \citep{mitchell2024}, but they do not remove the exponential number of coalition-specific alignments.

In this paper, we address this obstacle for Direct Alignment Algorithms (DAAs) \citep{rafailov2024}, a broad family of preference-based alignment objectives that includes DPO \citep{rafailov2023}, IPO \citep{azar2024}, and SLiC-style losses \cite{zhao2023}. Because these methods learn sequence-level log-policy ratios relative to a reference policy, aligning a model sequentially, one source at a time, exposes an additive structure. More precisely, under exact optimization at each stage, each source contributes a source-specific sequence-level log-ratio increment, and these increments add across updates. As a result, the score associated with any coalition can be reconstructed from the base model and the singleton models trained on individual sources, rather than by training a separate model for that coalition. This reduces the number of required alignments from exponential to linear in the number of sources. More generally, this reconstruction result is not specific to the Shapley value computation. It can be used whenever a downstream analysis requires complete-response scores for multiple coalitions of data sources. 

However, this exact reconstruction statement is idealized. In practice, neural training uses finite compute and a limited optimization budget\footnote{Throughout the paper, we use the terms \emph{finite training/finite optimization} to refer to this practical regime.}, so the alignment procedure may not reach the theoretical optimum at every stage. Whether coalition reconstruction remains accurate under these practical conditions is therefore an empirical question. We study this question directly across several base models and real-world preference sources. Specifically, we test whether sequential alignment is robust to the order in which sources are processed and whether coalition scores reconstructed from singleton models agree with those induced by policies trained sequentially on the same coalitions. In our main experiments, we instantiate the direct-alignment objective with IPO. We report complementary DPO results in Appendix~\ref{sec:additional-experiments}. We then compute Shapley values for these same sources under multiple reward models, showing how source valuations vary across evaluation criteria.

We summarize our contributions as follows.
\begin{itemize}

    \item[(a)] We propose a sequential preference optimization procedure for DAAs. 
    \item[(b)] We show that, in the idealized exact-optimization regime, the updates induced by sequential preference optimization compose additively in sequence-level log-policy-ratio space. Consequently, the log-probability assigned by any coalition policy to a fixed response can be reconstructed from the base policy and the policies trained on each source individually via a simple additive identity. This reduces the number of alignments required for Shapley-based valuation from exponential to linear in the number of sources.

    \item[(c)] We study whether this idealized reconstruction remains reliable under finite neural training. Across several base models and real-world preference sources, we test whether sequential alignment is robust to the order in which sources are processed and whether coalition scores reconstructed from singleton models agree with those induced by policies trained sequentially on the same coalitions. Our main experiments instantiate the direct-alignment objective with IPO, and we report complementary DPO results in Appendix~\ref{sec:additional-experiments}.

    \item[(d)] We apply the resulting reconstruction procedure to real-world preference sources and compute Shapley values for these same sources under multiple reward models, revealing how differently they contribute across evaluation criteria.
\end{itemize}

\paragraph{Related work.}
Two streams of Shapley value applications for aligned LLMs exist. One is focused on \emph{model explainability} for token- or feature-level attribution of predictions \citep{wang2024, xiao2025, horovicz2024, ye2025}, and another is focused on \emph{data valuation} for quantifying training data contributions \citep{schoch2023, he2024, wang2025, moon2025}. Our work is orthogonal to the first stream and contributes to the second. Its novelty, compared to prior data valuation studies for LLMs, is that it exploits the mathematical structure of preference-based alignment methods to reduce the computational cost of approximating coalition utilities for Shapley-based source valuation.

The computational cost of Shapley valuation has also motivated generic approximation methods that reduce the number of coalition utilities that must be evaluated, for instance through permutation Monte Carlo or other sampling schemes \citep{castro2009,maleki2013,ghorbani2019,jia2019b}. These methods reduce the number of coalitions considered, but in our setting each selected coalition would still require a separate alignment run. Our method targets this complementary source of cost by reducing alignment to $n$ trainings, regardless of the number of coalition utilities subsequently evaluated. Sampling-based Shapley approximations can therefore be combined with our reconstruction to further reduce the number of utility evaluations.

Our work is also related to methods that update the reference policy during offline preference optimization. Trust-region variants such as TR-DPO, TR-IPO, and TR-KTO update the reference policy within training on a single pooled dataset in order to stabilize optimization and mitigate overoptimization \citep{gorbatovski2024}. We also use a changing reference policy, but for a different purpose. Rather than improving a single pooled training run, we train sequentially across data sources in order to expose an additive structure in sequence-level log-policy-ratio space. We defer an extended discussion of related work to Appendix~\ref{sec:extended-related-work}.
\section{Preliminaries}
\label{sec:preliminaries}


\paragraph{Setup.} Let $\X$ be a prompt space and let $\Y$ be a finite set of complete responses. The finiteness assumption can be understood as fixing a finite vocabulary and a maximum response length. We represent a language model by a policy $\pi$ assigning a distribution $\pi(\cdot\mid x)$ over $\Y$ to every prompt $x\in\X$. We restrict attention to full-support policies and denote the base policy by $\pi_0$, with $\pi_0(y\mid x)>0$ for all $(x,y)\in\X\times\Y$, so that the log-policy ratios used in the paper are well defined.

We consider $n$ sources of preference data, indexed by $\N=\{1,\ldots,n\}$. Source $i$ is associated with a population distribution $P_i$ over preference triples $(x,y^+,y^-)$, where $y^+$ is preferred to $y^-$ for prompt $x$, and with a finite training sample $\D_i=\{(x_m,y_m^+,y_m^-)\}_{m=1}^{M_i}.$ A coalition is a subset $S\subseteq\N$. We write $\D_S=\bigcup_{i\in S}\D_i$ for the data provided by the sources in $S$.


\paragraph{Shapley-based data valuation.} Our goal is to quantify how much each preference source $i\in\N$ contributes to a chosen notion of alignment quality. A standard approach is Shapley-based data valuation \cite{ghorbani2019}, which treats each data source as a player in a cooperative game and assigns it a value based on its average marginal contribution across coalitions. To define such a game, we first associate each coalition $S\subseteq\N$ with a scalar utility $v(S)$ measuring the alignment quality obtained when the alignment procedure uses the sources in $S$. The empty coalition corresponds to the base model, so $v(\emptyset)$ measures the utility before preference alignment.

The definition of $v$ depends on the behavior one wishes to value. For example, a coalition may be evaluated by the average score of a reward model on a fixed set of prompts, or by the accuracy with which its complete-response scores rank held-out preferred responses above rejected responses. More generally, $v(S)$ is the scalar returned by a fixed evaluation protocol applied to the policy or response scores associated with coalition $S$. The same protocol is used for every coalition. Consequently, the resulting valuation is relative to the chosen utility rather than an intrinsic property of the data source itself \citep{tamine2025,diehl2025gameable}.

The Shapley value accounts for interactions between sources by averaging the marginal contribution of a source over all coalitions to which it can be added. For source $i\in\N$,
\begin{equation}
\label{eq:shapley}
\phi_i(v) =
\sum_{S\subseteq \N\setminus\{i\}}
\frac{|S|!\,(n-|S|-1)!}{n!}
\left[
v(S\cup\{i\})-v(S)
\right].
\end{equation}
One reason for the widespread use of the Shapley value in data valuation is that it satisfies a number of desirable axioms \citep{shapley1953}. Among them, the efficiency axiom requires that $\sum_{i\in\N}\phi_i(v) =v(\N)-v(\emptyset),$ meaning that the total change in utility obtained by using all preference sources, relative to the base model, is fully attributed among the sources.

However, directly implementing the Shapley value is prohibitively expensive for LLM alignment. Evaluating $v(S)$ would normally require training the model associated with each nonempty coalition and then applying the evaluation protocol. This requires $2^n-1$ alignment runs. Section~\ref{sec:efficient_shapley} shows how to avoid these coalition-specific trainings while retaining the quantities needed to evaluate Equation~\eqref{eq:shapley}.


\paragraph{Direct alignment algorithms.} In this paper, we use a Direct Alignment Algorithm (DAA) as the alignment procedure underlying the coalition utilities $v(S)$. A DAA optimizes a policy from preference comparisons through a loss applied to differences of complete-response log-policy ratios with respect to a reference policy. Formally, let $\pi_0$ be a reference policy satisfying
$\pi_0(y\mid x)>0$ for all $(x,y)\in\X\times\Y$ and define $h_{\pi,\pi_0}(x,y) = \log\frac{\pi(y\mid x)}{\pi_0(y\mid x)}.$ Given a convex loss $g$ and a scale parameter $\beta>0$, the population DAA
objective associated with source $i$ is
\begin{equation}
\label{eq:daa}
\mathcal L_i(\pi;\pi_0)
=
\mathbb E_{(x,y^+,y^-)\sim P_i}
\left[
g\left(
\beta
\left[
h_{\pi,\pi_0}(x,y^+)
-
h_{\pi,\pi_0}(x,y^-)
\right]
\right)
\right],
\end{equation}
where $\beta>0$ controls the scale of the reference-relative preference margin and $g:\mathbb R\to\mathbb R$ is a convex loss. DPO corresponds to $g(t)=-\log\sigma(t)$ \citep{rafailov2023}, IPO to $g(t)=(t-\tfrac12)^2$ up to a positive multiplicative constant \citep{azar2024}, and SLiC-style objectives correspond to hinge-type losses \citep{zhao2023}. In practice, we replace the expectation in Equation~\eqref{eq:daa} with its empirical average over $\D_i$.

The structure of Equation~\eqref{eq:daa} will be central in Section~\ref{sec:efficient_shapley}, where we show how to exploit it to evaluate coalitions of preference sources without training a separate model for each coalition.
\section{Efficient Shapley-based data valuation for LLM alignment}
\label{sec:efficient_shapley}

The computational bottleneck of Shapley-based data valuation is that it requires evaluating the utility of every coalition of data sources. If each coalition requires a separately aligned model, this amounts to $2^n-1$ alignment runs for $n$ sources. We show that, for DAAs and under exact population optimization, when alignment is performed sequentially over data sources, the information needed to evaluate any coalition can instead be recovered after training only one model per source. The proofs of all theoretical results in this section are given in Appendix~\ref{sec:missing-proofs}.


\subsection{Sequential preference optimization}
\label{subsec:sequential_training}

Let $S\subseteq\N$ be a coalition and let $\sigma=(\sigma_1,\ldots,\sigma_m)$ be an ordering of its $m=|S|$ sources. We propose a sequential preference optimization procedure that processes these sources one by one. Starting from $\pi^{(0)}=\pi_0$, stage $k$ optimizes the DAA objective \eqref{eq:daa} associated with source $\sigma_k$, using the current policy $\pi^{(k-1)}$ as reference:
\begin{equation}
\label{eq:sequential_update}
    \pi^{(k)}
    \in
    \argmin_{\pi}
    \mathcal L_{\sigma_k}
    \bigl(\pi;\pi^{(k-1)}\bigr),
    \qquad k=1,\ldots,m.
\end{equation}

In practice, Equation~\eqref{eq:sequential_update} is optimized on the finite dataset $\D_{\sigma_k}$. The theoretical analysis below instead considers the corresponding population objectives. For a singleton coalition $S=\{i\}$, the procedure consists of a single alignment on source $i$ starting from the base policy $\pi_0$. We denote its exact population counterpart by $\pi_i^\star$.

\subsection{Relative log-probability changes and order-independent alignment}
\label{subsec:exact_composition}
The only additional ingredient introduced by the sequential preference optimization procedure is that the reference policy changes from one source to the next. A priori, this could make the effect of processing a source depend on where that source appears in the sequence. We therefore first ask how the update induced by a source depends on the current reference policy.

The structure of the DAA objective makes this dependence particularly simple. Recall from Equation~\eqref{eq:daa} that, for a reference policy $\pi_0$, the candidate policy $\pi$ enters the objective through the complete-response log-policy ratio $h_{\pi,\pi_0}(x,y) = \log \frac{\pi(y\mid x)}{\pi_0(y\mid x)}.$ It is therefore convenient to reparameterize the optimization problem using the sequence-level log-policy ratio as a score coordinate, $f(x,y) = h_{\pi,\pi_0}(x,y).$ In these coordinates, the population objective associated with source $i$ becomes
\begin{equation}
\label{eq:population_score_objective}
\E_{(x,y^+,y^-)\sim P_i}
\left[
g\left(
\beta
\left[
f(x,y^+)-f(x,y^-)
\right]
\right)
\right].
\end{equation}

Equation~\eqref{eq:population_score_objective} makes explicit that the objective depends on $f$ only through differences between scores assigned to responses to the same prompt. It is therefore unchanged when an arbitrary prompt-dependent constant $c(x)$ is added to $f(x,\cdot)$.

We formalize the ideal regime in which these score differences are well-defined.

\begin{assumption}[Exact and identifiable population optimum]
\label{ass:exact_optimization}
For every source $i$, the population objective \eqref{eq:population_score_objective} attains a finite minimum, and its minimizers are unique up to a prompt-dependent additive
constant. Formally, for any two minimizers $f$ and $\widetilde f$, there exists
$c:\X\to\mathbb R$ such that $\widetilde f(x,y)=f(x,y)+c(x)$ for all $(x,y)\in\X\times\Y$. At every stage of the theoretical sequential procedure, the corresponding population minimum is attained.
\end{assumption}

This assumption idealizes the optimization regime. Section~\ref{sec:finite-training-validation} assesses how well the analysis developed below in this idealized regime
applies to finite neural training.

Consider now source $i \in \mathcal{N}$ and an arbitrary reference policy $\pi$ satisfying $\pi(y\mid x)>0$ for all $(x,y)$ and define
\begin{equation*}
T_i^\star(\pi)(y\mid x)
=
\frac{
\pi(y\mid x)\exp\!\left(f_i^\star(x,y)\right)
}{
\sum_{y'\in\Y}
\pi(y'\mid x)\exp\!\left(f_i^\star(x,y')\right)
}.
\end{equation*}

\begin{lemma}[Reference independence of relative log-probability changes]
\label{lem:reference_invariance}
Under Assumption~\ref{ass:exact_optimization}, $T_i^\star(\pi)$ is an exact
population optimizer for source $i$ with reference $\pi$. Moreover, for any
prompt $x$ and any two responses $y,y'\in\Y$,
\begin{equation}
\label{eq:reference_invariance}
\log
\frac{T_i^\star(\pi)(y\mid x)}{\pi(y\mid x)}
-
\log
\frac{T_i^\star(\pi)(y'\mid x)}{\pi(y'\mid x)}
=
f_i^\star(x,y)-f_i^\star(x,y').
\end{equation}
The right-hand side depends only on source $i$ and is the same for every
full-support reference policy $\pi$. Hence, the change induced by source $i$
in the relative log-probability of two responses does not depend on the
reference policy.
\end{lemma}
Thus, under exact population optimization, training on source $i$ changes the relative log-probability of any two responses by an amount determined only by that source, independently of the reference policy from which optimization starts. This immediately yields order-invariant composition.

\begin{proposition}[Order-invariant composition across sources]
\label{prop:composition}
Let $S\subseteq\N$ and let $\sigma$ be any ordering of the sources in $S$. Under Assumption~\ref{ass:exact_optimization}, the sequential preference optimization procedure \eqref{eq:sequential_update} returns
\begin{equation*}
\pi_S^\star(y\mid x)
=
\frac{
\pi_0(y\mid x)
\exp\!\left(
\sum_{i\in S} f_i^\star(x,y)
\right)
}{
\sum_{y'\in\Y}
\pi_0(y'\mid x)
\exp\!\left(
\sum_{i\in S} f_i^\star(x,y')
\right)
}.
\end{equation*}
Consequently, the resulting policy depends only on the coalition $S$ and not on the ordering $\sigma$.
\end{proposition}

\subsection{Reconstruction from singleton models}
\label{subsec:reconstruction}

Proposition~\ref{prop:composition} characterizes the coalition policy through the source scores $f_i^\star$. Crucially, these scores do not need to be estimated separately. They are already encoded, up to prompt-dependent
constants, in the policies obtained by aligning the base model on a single
source. We refer to these as \emph{singleton policies}. Let $\pi_i^\star = T_i^\star(\pi_0)$ be the exact policy obtained by optimizing on source $i$ alone from the base policy. From Lemma~\ref{lem:reference_invariance},
\begin{equation*}
\log\pi_i^\star(y\mid x)
-
\log\pi_0(y\mid x)
=
f_i^\star(x,y)-c_i(x),
\end{equation*}
where $c_i(x)$ does not depend on the response $y$.

For any coalition $S\subseteq\N$, define
\begin{equation*}
\begin{split}
\widehat s_S^\star(x,y)
&=
\log\pi_0(y\mid x)
+
\sum_{i\in S}
\left[
\log\pi_i^\star(y\mid x)
-
\log\pi_0(y\mid x)
\right]
\\
&=
\sum_{i\in S}\log\pi_i^\star(y\mid x)
+
(1-|S|)\log\pi_0(y\mid x).
\end{split}
\end{equation*}
For $S=\emptyset$, this definition gives $\widehat s_\emptyset^\star(x,y)=\log\pi_0(y\mid x)$.

\begin{proposition}[Exact reconstruction from singleton models]
\label{prop:reconstruction}
Under Assumption~\ref{ass:exact_optimization}, for every coalition $S\subseteq\N$ there exists a function $c_S:\X\to\mathbb{R}$ such that
\begin{equation*}
\widehat s_S^\star(x,y)
=
\log\pi_S^\star(y\mid x)+c_S(x).
\end{equation*}
Consequently, for any two responses $y,y'$ to the same prompt $x$,
\begin{equation*}
\log\pi_S^\star(y\mid x)
-
\log\pi_S^\star(y'\mid x) =
\widehat s_S^\star(x,y)
-
\widehat s_S^\star(x,y').
\end{equation*}
\end{proposition}

Proposition~ \ref{prop:reconstruction} is the central result of this paper. It says that the singleton policies recover any coalition log-policy up to a constant that depends on the prompt but not on the candidate response. This unidentified constant disappears whenever responses are compared or normalized within the same prompt. In particular, for any fixed nonempty candidate set $\mathcal C(x)\subseteq\Y$ and any
$y\in\mathcal C(x)$,
\begin{equation*}
\frac{
\exp\!\left(\widehat s_S^\star(x,y)\right)
}{
\sum_{y'\in\mathcal C(x)}
\exp\!\left(\widehat s_S^\star(x,y')\right)
}
=
\frac{
\pi_S^\star(y\mid x)
}{
\sum_{y'\in\mathcal C(x)}
\pi_S^\star(y'\mid x)
}.
\end{equation*}
Hence, the reconstruction exactly preserves pairwise score differences, rankings, argmax decisions, and normalized distributions over any fixed candidate set.

\subsection{\texorpdfstring{Shapley-based valuation with $n$ alignments}{Shapley-based valuation with n alignments}}
\label{subsec:shapley_from_singletons}

We can now return to the data valuation problem introduced in
Section~\ref{sec:preliminaries}. Let $\mathcal V:
\mathbb R^{\mathcal X\times\mathcal Y}
\to \mathbb R$ denote the fixed
evaluation protocol used to define coalition utility. We require only that
$\mathcal V$ be invariant to adding a prompt-dependent constant to all
response scores for that prompt, i.e.,
\begin{equation}
\label{eq:shift_invariant_utility}
\mathcal V(s+c)=\mathcal V(s)
\end{equation}
for any function $c(\cdot)$ that does not depend on the response. 

In the exact population regime, we define the coalition utility by $v(S)=
\mathcal V\!\left(\log\pi_S^\star\right).$
Proposition~\ref{prop:reconstruction} and
Equation~\eqref{eq:shift_invariant_utility} imply
$v(S) = \mathcal V\!\left(\widehat s_S^\star\right).$
Thus, every coalition utility can be recovered from the base policy and the
singleton policies, without training a model on that coalition.

The corresponding Shapley value of source $i$ is therefore
\begin{equation*}
\phi_i(v)
=
\sum_{S\subseteq\N\setminus\{i\}}
\frac{|S|!\,(n-|S|-1)!}{n!}
\left[
\mathcal V\!\left(\widehat s_{S\cup\{i\}}^\star\right)
-
\mathcal V\!\left(\widehat s_S^\star\right)
\right].
\end{equation*}

The computational cost of Shapley valuation has two distinct components: training the models associated with coalitions and evaluating their utilities. Our method reduces the first, and typically much more expensive, component. A direct implementation requires training one aligned model for each of the $2^n-1$ nonempty coalitions, whereas our reconstruction requires only the $n$ singleton policies $\pi_1^\star,\ldots,\pi_n^\star$. Exact Shapley computation still requires evaluating the utilities of all $2^n$ coalitions, but these utilities can now be obtained from reconstructed coalition scores without any additional alignment. Our method can therefore be combined with standard Shapley approximation techniques, such as Monte Carlo permutation sampling, to also reduce the number of coalition utilities that must be evaluated when exhaustive enumeration is too costly \citep{castro2009,maleki2013,ghorbani2019}.

In practice, the singleton policies are obtained by finite neural optimization rather than exact population optimization. Given the learned singleton models $\pi_1,\ldots,\pi_n$, we therefore use
\begin{equation}
\label{eq:practical_reconstruction}
\widehat s_S(x,y)
=
\log\pi_0(y\mid x)
+
\sum_{i\in S}
\left[
\log\pi_i(y\mid x)-\log\pi_0(y\mid x)
\right]
\end{equation}
and define the reconstructed coalition utility $\widehat v(S) = \mathcal V\!\left(\widehat s_S\right).$

Unlike under exact population optimization,
Equation~\eqref{eq:practical_reconstruction} is not guaranteed to match the complete-response scores of a coalition obtained by sequentially training on
sources in it. Section~\ref{sec:finite-training-validation} therefore empirically tests whether the properties established in the idealized regime
remain approximately valid in practice.
Once we assess this empirical approximation, Section~\ref{sec:shapley_application} instantiates $\mathcal V$ with alignment criteria and uses $\widehat v$ to compute Shapley values for real-world data sources.
\section{Finite-training validation}
\label{sec:finite-training-validation}

Section~\ref{sec:efficient_shapley} establishes two consequences of sequential preference alignment under exact population optimization that are central to our method. First, sequential optimization over a coalition is independent of the order in which its sources are processed. Second, the behavior of any coalition can be reconstructed at the level of complete-response scores using only the policies trained on individual sources. However, neither property is guaranteed to hold exactly after finite neural-network training. We therefore test these two consequences directly. 

\subsection{Experimental protocol}
\label{subsec:finite_training_protocol}

We construct $n=9$ preference sources from the UltraFeedback dataset \citep{cui2023ultrafeedback} by partitioning prompts according to their original data provenance. For each source $i$, the dataset $\D_i$ contains preference triples $(x,y^+,y^-)$ from that provenance, where $y^+$ and $y^-$ are the highest- and lowest-scoring valid completions available for prompt $x$, respectively. 

We evaluate finite-training behavior across SmolLM \citep{allal2024smollm}, Qwen2.5 \citep{qwen2025qwen25},
Llama3.2 \citep{grattafiori2024llama} and Phi-3.5 \citep{abdin2024phi} models. All models are aligned with IPO \citep{azar2024}. Full experimental details are provided in Appendix~\ref{sec:xp-details}.

We consider coalitions of sizes $|S|\in\{3,6,9\}$. For each evaluated coalition, we specify three source orderings $\Sigma_S=\{\sigma_1,\sigma_2,\sigma_3\}.$ For an ordering $\sigma\in\Sigma_S$, let $\pi_{S,\sigma}$ denote the policy obtained by the sequential preference optimization procedure specified in Section~\ref{subsec:sequential_training}. We train different orders under the same seed and identical experimental conditions. We also repeat the same ordered training with independent optimization seeds to estimate variability due to finite training. Appendix~\ref{sec:xp-details} lists the exact coalitions, source orders, and repetitions.

All comparisons are performed on the same held-out evaluation set $\mathcal E =  \{(x_j,y_j^+,y_j^-)\}_{j=1}^{M}$.

\subsection{A common discrepancy measure}
\label{subsec:finite_training_metrics}

Proposition~\ref{prop:reconstruction} shows that $\widehat s_S^\star(x,y)$ recovers $\log\pi_S^\star(y\mid x)$ up to a prompt-dependent additive constant. Because this constant is the same for all responses to a given prompt, it cancels when taking the difference between the scores assigned to two responses. We therefore evaluate finite-training deviations through such pairwise differences.

For a policy $\pi$, define its preference gap on evaluation example $(x_j,y_j^+,y_j^-) \in \mathcal{E}$ as $G_\pi^{(j)} = \log\pi(y_j^+\mid x_j) - \log\pi(y_j^-\mid x_j)$. Given two vectors of preference gaps $G$ and $G'$, we measure their discrepancy
by
\begin{equation*}
d(G,G')
=
\frac{1}{M}
\sum_{j=1}^{M}
\left|
G^{(j)}-G'^{(j)}
\right|.
\end{equation*}

However, raw values of $d$ are not directly comparable across models and coalition sizes, since the scale of finite-training variability itself may differ. To obtain a comparable reference scale, for each considered coalition $S$, we fix one reference ordering $\sigma_{\mathrm{ref}}$. We train this same ordered coalition $R$ times, using different optimization seeds $r\in[R]$, while keeping the data, source order, and all other training choices fixed. Let $\pi_{S,\sigma_{\mathrm{ref}}}^{(r)}$ denote the policy obtained in run $r$. We define the
\emph{retraining floor\footnote{We use \emph{floor} as shorthand for this empirical reference scale. It is not assumed to be a theoretical lower bound.}} as the average pairwise discrepancy between these $R$ runs, i.e.,
\begin{equation*}
d_{\mathrm{floor}}(S)
=
\frac{2}{R(R-1)}
\sum_{1\leq r<r'\leq R}
d\!\left(
G_{\pi_{S,\sigma_{\mathrm{ref}}}^{(r)}},
G_{\pi_{S,\sigma_{\mathrm{ref}}}^{(r')}}
\right).
\end{equation*}

Both experiments below report their discrepancy relative to this same quantity. 

\subsection{Robustness to source order}
\label{subsec:order_robustness}

Proposition~\ref{prop:composition} predicts that, under exact population optimization, the final coalition policy does not depend on the source ordering. Under finite training, we measure how far this order invariance
degrades. For two distinct orderings $\sigma,\tau\in\Sigma_S$, we define the order discrepancy as $d_{\mathrm{order}}(S;\sigma,\tau) = d\!\left(G_{\pi_{S,\sigma}},G_{\pi_{S,\tau}}\right),$ and normalize it by the retraining floor to obtain 
\begin{align*}
\rho_{\mathrm{order}}(S;\sigma,\tau) = \frac{d_{\mathrm{order}}(S;\sigma,\tau)}{d_{\mathrm{floor}}(S)}.
\end{align*}

$\rho_{\mathrm{order}}=1$ means that changing the source order causes a discrepancy equal to ordinary retraining variability.

\subsection{Reconstruction fidelity}
\label{subsec:reconstruction_fidelity}

We next test Proposition~\ref{prop:reconstruction} under finite training. Equation~\eqref{eq:practical_reconstruction} assigns a reconstructed score
$\widehat s_S(x,y)$ to each response. We define the corresponding reconstructed
preference gap on evaluation example $(x_j,y_j^+,y_j^-) \in \mathcal{E}$ as $\widehat G_S^{(j)} = \widehat s_S(x_j,y_j^+) - \widehat s_S(x_j,y_j^-).$

For each ordering $\sigma$, we compare this reconstructed gap with the gap of the corresponding sequentially trained policy $d_{\mathrm{recon}}(S;\sigma)
=
d\!\left(
\widehat G_S,
G_{\pi_{S,\sigma}}
\right).$
We again normalize by the same retraining floor,
\begin{equation*}
\rho_{\mathrm{recon}}(S;\sigma)
=
\frac{
d_{\mathrm{recon}}(S;\sigma)
}{
d_{\mathrm{floor}}(S)
}.
\end{equation*}

$\rho_{\mathrm{recon}}=1$ means that the reconstruction discrepancy is equal to the retraining variability measured under the fixed reference ordering.

\subsection{Results}
\label{subsec:finite_training_results}

\begin{figure*}[h!]
    \centering

    \begin{subfigure}[t]{0.49\textwidth}
        \centering
        \includegraphics[width=\linewidth]{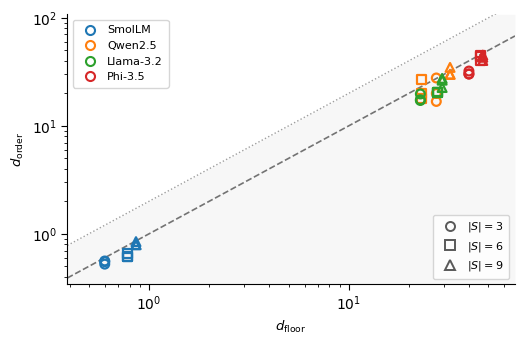}
        \caption{Order robustness}
        \label{fig:order_robustness}
    \end{subfigure}
    \hfill
    \begin{subfigure}[t]{0.49\textwidth}
        \centering
        \includegraphics[width=\linewidth]{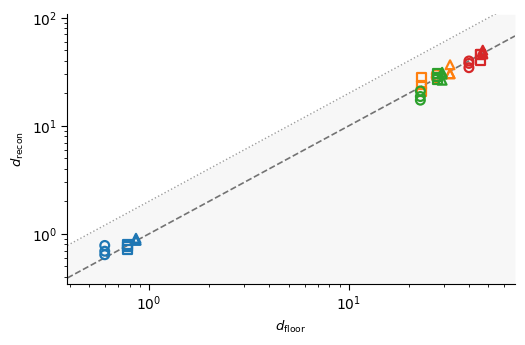}
        \caption{Reconstruction fidelity}
        \label{fig:reconstruction_fidelity}
    \end{subfigure}

    \caption{
    Finite-training validation of the two properties underlying efficient Shapley-based valuation for LLM alignment. Each point compares a measured discrepancy (vertical axis) with the retraining floor for the corresponding model and coalition (horizontal axis). Left: $d_{\mathrm{order}}$, the discrepancy between sequential trainings that differ in source order. Right: $d_{\mathrm{recon}}$, the discrepancy between reconstructed coalition gaps and an explicitly trained sequential coalition. The dashed diagonal corresponds to $\rho=1$. The shaded region indicates $y\leq 2x$, i.e., discrepancies no larger than twice the retraining floor. Colors identify the four base models and marker shapes indicate coalition size $|S|\in\{3,6,9\}$. Three prespecified source orderings are evaluated for each coalition size.}
    \label{fig:finite_training_validation}
\end{figure*}

Figure~\ref{fig:finite_training_validation} summarizes both tests using the
same representation. 

Across all evaluated models, coalition sizes, and source orderings, both order and reconstruction discrepancies remain within twice the retraining floor. Thus, in all tested configurations, changing the source order or reconstructing a coalition from singleton policies introduces deviations that remain on the same scale as the variability induced by finite training itself. These results do not imply exact order invariance or exact reconstruction under finite optimization, but they show that the departures from the ideal theory remain controlled relative to retraining variability.
\section{Application to Shapley-based Data Valuation}
\label{sec:shapley_application}

We finally apply the reconstruction method to estimate the contribution of the
$n=9$ preference sources. Since Section~\ref{sec:finite-training-validation} showed that
finite-training errors remain comparable to ordinary
retraining variability, we use the learned singleton policies to evaluate all
coalitions without additional alignment.

Let $\mathcal E = \{(x_j,y_j^+,y_j^-)\}_{j=1}^{M}$
be the held-out evaluation set introduced in
Section~\ref{sec:finite-training-validation}, and let
$\mathcal X_{\mathcal E} = \{x_j\}_{j=1}^{M}$
denote its prompts. For each $x \in \mathcal X_{\mathcal E}$, we generate once a fixed candidate set $\mathcal C(x)=\{y_1,\ldots,y_K\}$ from a coalition-independent proposal
policy. We use the same candidates for every coalition. For
$S\subseteq\N$, we reconstruct their scores using
Equation~\eqref{eq:practical_reconstruction} and select
\begin{equation}
\label{eq:best_of_k}
\widehat y_S(x)
=
\arg\max_{y\in\mathcal C(x)}
\widehat s_S(x,y).
\end{equation}
Under exact population optimization,
Proposition~\ref{prop:reconstruction} implies that
$\widehat s_S^\star(x,y)$ differs from
$\log\pi_S^\star(y\mid x)$ only by a prompt-dependent constant. Hence,
Equation~\eqref{eq:best_of_k} selects the same response as the exact coalition
policy within the fixed candidate set.

We evaluate the selected responses with three reward models, namely Skywork-Reward-V2
\citep{liu2026skywork}, GRM \citep{yang2024grm}, and UltraRM \citep{cui2023ultrafeedback}, respectively denoted as $\mathcal R=\{r_1,r_2,r_3\}$ where each
$r:\X\times\Y\to\mathbb R$ assigns a scalar reward $r(x,y)$ to response $y$
for prompt $x$. Because their numerical scales may differ and to make the comparison across
reward models visually interpretable, each reward is standardized over the common set of evaluated prompt-candidate
pairs, i.e., $\widetilde r(x,y)=
\frac{r(x,y)-\mu_r}{\sigma_r}$ where $\mu_r$ and $\sigma_r$ denote the mean and standard deviation of reward $r$ over all
$x\in\mathcal X_E$ and $y\in C(x)$. The utility of coalition $S$ under reward $r$ is then  $\widehat v_r(S)
=
\frac{1}{|\mathcal X_{\mathcal{E}}|}
\sum_{x\in\mathcal X_{\mathcal{E}}}
\widetilde r\!\left(x,\widehat y_S(x)\right).$


\begin{wrapfigure}{r}{0.6\textwidth}
    \centering
    \vspace{-0.8em}
    \includegraphics[width=\linewidth]{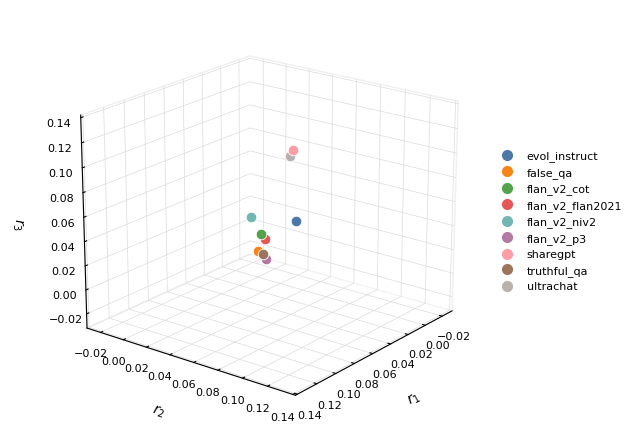}
    \caption{
    Shapley values of the $n=9$ preference sources for Qwen2.5 under three standardized
    reward models. Each source $i$ is represented by
    $(\phi_{i}(\widehat{v}_{r_1}),
      \phi_{i}(\widehat{v}_{r_2}),
      \phi_{i}(\widehat{v}_{r_3})$.
      Along axis $r_k$, coordinates are expressed in standard-deviation units of the corresponding reward score.
    }
    \label{fig:shapley_3d}
    \vspace{-0.5em}
\end{wrapfigure}

For each reward $r$ and each source $i$, we compute the Shapley value $\phi_i(\widehat{v}_r)$ from the reconstructed coalition utilities\footnote{Full experimental details are given in Appendix~\ref{sec:xp-details}.}. Figure~\ref{fig:shapley_3d} shows each source's Shapley value when the base model is Qwen2.5, under the three reward models. Equivalent results for SmolLM, Llama3.2 and Phi-3.5 (the other models evaluated in Section~\ref{sec:finite-training-validation}) are reported in Appendix~\ref{sec:additional-figures}.

The resulting valuations are broadly consistent across reward models. Sources receiving a large positive contribution under one reward tend to remain positively valued under others, while several sources cluster near the origin and therefore contribute only marginally to the corresponding games. ShareGPT and Ultrachat are separated from the remaining sources along the $r_3$ axis. By contrast, most \texttt{flan}-derived sources lie closer to the origin.

\wrapfill
More generally, our method makes Shapley-based valuation tractable, providing an interpretable view of which data sources contribute to the aligned model's performance and how these contributions depend on the chosen reward.

\section{Conclusion}
\label{sec:conclusion}

We introduced a method for Shapley-based valuation of preference data sources that avoids training a separate model for every coalition. Under exact population optimization, we showed that sequentially optimizing a DAA over the sources in a coalition gives the same final policy regardless of their order. We further showed that, for any complete response, the log-probability assigned by this coalition policy can be recovered up to a prompt-dependent constant from the base policy and the policies trained on each source individually. As a result, evaluating coalitions requires only $n$ singleton alignments instead of $2^n-1$
coalition-specific alignments. We then tested whether these theoretical properties persist under finite training. Across several models, changing the source order and reconstructing coalition scores from singleton policies both introduced discrepancies no larger than twice the retraining variability. Finally, we applied our method to $n=9$ real-world preference sources
and computed their Shapley values under three reward models, a computation that would otherwise require $2^9 - 1 = 511$ alignment runs and be prohibitively expensive in practice.
\paragraph{Limitations.} Our reconstruction operates at the level of sequence log-probabilities for fixed complete responses. It does not reconstruct a generative coalition policy. Appendix~\ref{subsec:next-token-gen-discussion} explains why obtaining next-token probabilities would additionally require marginalization over future continuations. Extending the method to support direct generation is a natural direction for future work.
\newpage
\subsection*{AI use statement}

In this work, we used generative AI tools to aid or polish writing and to assist with code generation. All authors carefully reviewed the writing. AI-assisted code was reviewed and checked by at least one author. We take responsibility for the final content of this work, including text, claims, or artifacts produced with the aid of generative AI.

\subsection*{Reproducibility statement}

An anonymized code package is provided in the supplementary material. It implements the data preparation, alignment, evaluation, valuation,
and plotting procedures used in the paper. Appendix \ref{sec:xp-details} documents full experimental protocols, including datasets, hyperparameters, and compute settings, and cross-references them at the relevant points in the main text. Appendix \ref{sec:missing-proofs} contains all missing proofs of theoretical results in the main text.
\subsubsection*{Acknowledgments}
This work was partially supported by the French National Research Agency (ANR) through grants ANR-20-CE23-0007 and ANR-23-CE23-0002 and through the PEPR IA FOUNDRY project (ANR-23-PEIA-0003). This work benefited from Hi! PARIS and State funding managed by the French National Research Agency (ANR) under the France 2030 program, reference ANR-23-IACL-0005. Computational and storage resources were provided by GENCI at IDRIS through allocation 2025-A0191016862 on the Jean Zay supercomputer (V100/A100/H100 partitions).

\bibliographystyle{iclr2027_conference}
\bibliography{bibliography}

\newpage
\appendix
\paragraph{Outline.} The appendices are organized as follows.
\begin{itemize}
    \item[--] Appendix~\ref{sec:extended-related-work} provides an extended discussion of related work.
    \item[--] Appendix~\ref{sec:missing-proofs} contains the proofs omitted from the main text.
    \item[--] Appendix~\ref{sec:xp-details} provides experimental details for all empirical results reported in the main paper.
    \item[--] Appendix~\ref{sec:additional-experiments} reports additional experimental results.
    \item[--] Appendix~\ref{sec:additional-figures} provides additional figures complementing the results presented in the main paper.
    \item[--] Appendix~\ref{sec:additional-discussions} contains additional discussions and technical remarks.
\end{itemize}

\section{Extended related work}
\label{sec:extended-related-work}

\subsection{Shapley-based data valuation}
\label{subsec:rw-shapley-data-valuation}

Shapley-based data valuation has become popular in recent years because it provides an axiomatic framework for attributing a trained model's performance to the data used to train it. In this framework, each training example is treated as a player in a cooperative game. The utility of a coalition is the performance of a model trained on the corresponding subset, and the Shapley value \cite{shapley1953} assigns each example its average marginal contribution over coalitions \citep{ghorbani2019,jia2019b}. This formulation has been used for data debugging, data selection, noisy-label detection, and dataset pricing \citep{sim2022,jiang2023}. The same cooperative-game principle extends from data points to datasets or sources: each contributor provides a dataset, and the goal is to value each source's contribution to the final model. This source-level formulation is the one considered in our paper.

The appeal of the Shapley value comes from its axiomatic characterization. It is the unique valuation rule satisfying four key axioms:
\begin{enumerate}
    \item \emph{Null player.} If a player has zero marginal contribution to every coalition, i.e., $u(S\cup\{i\}) = u(S) \forall S\subseteq \mathcal I\setminus\{i\},$ then its value is zero: $\phi_i(u)=0$.
    \item \emph{Symmetry.} If two players have identical marginal contributions to every coalition, i.e., $u(S\cup\{i\}) = u(S\cup\{j\}) \forall S\subseteq \mathcal I\setminus\{i,j\},$ then they receive the same value: $\phi_i(u)=\phi_j(u)$.
    \item \emph{Efficiency.} The total value is distributed among all players: $\sum_{i\in\mathcal I}\phi_i(u)=u(\mathcal I)-u(\emptyset)$.
    \item \emph{Additivity.} For any two games $u$ and $w$ on the same player set, $\phi_i(u+w)=\phi_i(u)+\phi_i(w),\forall i\in\mathcal I$.
\end{enumerate}

A growing body of work applies Shapley-based data valuation to language model fine-tuning and instruction tuning. \cite{schoch2023} proposes transferred Shapley values for data selection in LLM fine-tuning, using value information from simpler models to reduce valuation costs. \cite{he2024} introduces SHED, a Shapley-based automated dataset refinement framework for instruction fine-tuning. \cite{moon2025} proposes LimaCost, a data valuation method for instruction tuning that estimates an example's value by the amount of reference data needed to approximate its gradient. These methods share our goal of identifying valuable data for LLM adaptation,
but they do not exploit the algebraic structure of preference-optimization
objectives to reconstruct complete-response coalition scores from singleton
models.

Finally, several works study Shapley values for LLMs from an explainability rather than a training-data valuation perspective. Token-level methods such as TokenSHAP and TokenShapley attribute model outputs to prompt or context tokens \citep{horovicz2024,xiao2025}. Other work studies document-level attribution in generated summaries using Shapley-style document valuation \citep{ye2025}. These methods explain a prediction, a generated answer, or a summary. Our work instead values preference datasets used during alignment: the players are data sources, and the utility of a coalition is the behavior of the aligned model obtained from those sources.

\subsection{Efficient Shapley value estimation methods}
\label{subsec:rw-efficient-shapley}

The main computational obstacle in Shapley-based data valuation is the exponential number of coalitions. For $n$ players, exact computation requires evaluating utilities for all $2^n$ coalitions. Classical approximation methods, therefore, estimate the Shapley value by sampling coalitions or permutations. Permutation Monte Carlo estimates the Shapley value by averaging marginal contributions along random permutations of players \citep{castro2009,maleki2013,ghorbani2019}. Stratified sampling improves this approach by allocating samples across coalition sizes \citep{maleki2013}. Group-testing methods estimate the Shapley vector via randomized coalition queries and pairwise-difference constraints \citep{jia2019b}. These methods are generic: they reduce the number of coalitions queried, but each queried coalition must still be evaluated by running the learning procedure associated with it.

Other works exploit structure in specific learning algorithms. For nearest-neighbor classifiers, \cite{jia2019a} derives efficient task-specific data valuation procedures that avoid generic retraining over coalitions. More recently, \cite{wang2025} show how Data Shapley values can be estimated
from a single training run by accumulating contribution estimates along the
training trajectory. These methods show that Shapley computation can become practical when one uses the training algorithm's structure rather than treating the learning process as a black-box utility oracle.

Our work follows the same high-level philosophy (using algorithmic structure to reduce valuation cost) but in a different way. We consider source-level valuation for preference-based LLM alignment. Rather than directly approximating the Shapley summation, we reduce the cost of evaluating coalition utilities. Under the exact-optimization structure of Direct Alignment Algorithms,
source-level updates compose additively in sequence-level log-policy-ratio
space. This makes it possible to train one singleton model per source and
reconstruct the complete-response scores needed to evaluate coalition
utilities, without coalition-specific alignment. Thus, our method is complementary to Monte Carlo or group-testing estimators: once coalition utilities can be cheaply queried via reconstruction, standard Shapley approximation methods can still be used to reduce the number of coalition utility evaluations.

\subsection{Semivalue-based data valuation}
\label{subsec:semivalue-based-data-valuation}
The Shapley value is one member of the broader family of semivalues \citep{dubey1981}. Semivalues assign value through weighted averages of marginal contributions, with coalitions of the same size receiving the same weight, while allowing the weighting across coalition sizes to differ from
that of the Shapley value. In particular, they need not satisfy the efficiency property that uniquely characterizes the Shapley weighting within this family.

This additional flexibility has been studied for data valuation. \citet{kwon2021beta} introduce Beta Shapley, which uses a family of cardinality-dependent weights to reduce the effect of noisy marginal contributions and contains Data Shapley as a special case. \cite{wang2023data} study the Banzhaf value as a data valuation rule and show that its alternative coalition weighting can provide greater robustness to stochastic utility evaluations. These works show that the choice of weights across coalition sizes can be useful when properties beyond Shapley efficiency are desired.

A related line of work studies the sensitivity of semivalue-based data valuation to the definition or estimation of the underlying utility. \cite{tamine2025} analyzes how changes in the utility function affect semivalue-based rankings, while \cite{diehl2025gameable} studies the extent
to which such valuations can depend on an underspecified utility. These results emphasize that, for Shapley values as well as other semivalues, data values are defined relative to both a coalition utility and a choice of marginal-contribution weighting. We discuss in Appendix~\ref{subsec:semivalue_extension} how our coalition reconstruction can
be used with arbitrary semivalue weights.

\subsection{Alternative data valuation frameworks}
\label{subsec:rw-alternative-data-valuation}

Beyond Shapley-based methods, and more generally semivalue-based methods, several approaches assign value or influence scores to data without using the exact cooperative-game formulation. Influence functions approximate the effect of upweighting or removing a training point through derivatives of the learned parameters \citep{koh2017}. TracIn and related methods estimate data influence by accumulating gradient similarities along optimization trajectories \citep{pruthi2020tracin}. Representer-point methods decompose predictions through training examples in representation space \citep{yeh2018representer}. Data models learn a predictive model of how training subsets affect model outputs, providing a flexible empirical representation of data influence \citep{ilyas2022datamodels}. These methods are often more scalable than exact Shapley computation, but they generally define different notions of influence rather than the Shapley value itself.

In the LLM setting, data selection and data attribution have also been studied through non-Shapley criteria. Some methods select instruction-tuning examples based on diversity, quality scores, gradient similarity, or influence-style quantities. These approaches are useful for improving fine-tuning efficiency, but they typically do not preserve the cooperative-game interpretation of marginal contribution across data sources. By contrast, our objective is to keep the Shapley value as the valuation rule and make its coalition utilities tractable for preference-aligned LLMs.

A related robustness line of research examines how sensitive semivalue-based data valuations are to modeling choices. \cite{tamine2025} shows that changing the utility function can substantially alter semivalue-based rankings and introduces tools for analyzing utility-dependent robustness. \cite{diehl2025gameable} argue that semivalue-based data valuation can be arbitrary or gameable when the utility is underspecified. These works highlight that data values are meaningful only relative to a clearly specified utility. 

\subsection{Language model arithmetic and model merging}
\label{subsec:rw-lm-arithmetic-model-merging}

Our reconstruction rule is closely related to recent work showing that pretrained or fine-tuned models can often be combined algebraically. 
Language model arithmetic composes language models at inference time by
combining their output distributions or logits, enabling controlled text
generation without additional fine-tuning \citep{dekoninck2024}. Our method is related
in that it also combines information encoded by independently aligned models,
but the object being reconstructed is different. Our theoretical result is
derived at the level of complete-response log-probabilities: under exact
population optimization, the score assigned by a coalition policy to any fixed
complete response can be recovered, up to a prompt-dependent additive
constant, from the base policy and the corresponding singleton policies. In
practice, we use this identity only to score and compare fixed complete
responses. We do not locally combine next-token distributions to define a
generative coalition policy. Appendix~\ref{sec:additional-discussions}
discusses this distinction in detail.

Model merging and task arithmetic provide another related perspective. Model soups average weights of fine-tuned models and can improve accuracy without increasing inference cost \citep{wortsman2022model}. Task arithmetic represents fine-tuning effects as task vectors in weight space and combines these vectors to edit model behavior \citep{ilharco2023task}. TIES-merging studies interference between task vectors and proposes sign-resolution and trimming steps to merge models more robustly \citep{yadav2023ties}. These methods demonstrate that independently fine-tuned models can sometimes be combined without retraining. However, they generally operate in parameter space and are motivated by
multi-task adaptation or model editing. Our method instead operates in
complete-response log-probability space and is tied specifically to preference
optimization: the composition rule reconstructs the complete-response scores
associated with a coalition from the base and singleton policies, enabling
Shapley utilities to be evaluated without coalition-specific alignment.

Our work is also related to methods that update the reference policy during offline alignment. Trust-region variants such as TR-DPO, TR-IPO, and TR-KTO update the reference policy during training to improve stability and mitigate overoptimization \citep{gorbatovski2024}. Sequential Preference Optimization also updates the reference policy, but for a different purpose. Trust-region methods use reference updates within a single training run on a pooled dataset. We use source-wise reference updates to expose an additive structure across data sources. This distinction is important: our sequential procedure is not primarily an
optimization trick, but a way to define the alignment procedure associated
with each coalition so that, under the assumptions of our theory, its
complete-response scores can be reconstructed from singleton policies and used
for Shapley-based data valuation.
\newpage

\section{Missing proofs}
\label{sec:missing-proofs}

This appendix provides complete proofs of the theoretical results stated in
Section~\ref{sec:efficient_shapley}. Throughout, we work under
Assumption~\ref{ass:exact_optimization} and with full-support policies, as in
the main text.

\subsection{Proof of Lemma~\ref{lem:reference_invariance}}

\begin{lemma}[Restatement of Lemma \ref{lem:reference_invariance}]
Fix a source $i\in\N$, let $f_i^\star$ be any minimizer of the population
score objective~\eqref{eq:population_score_objective}, and let $\pi$ be any
full-support reference policy. Define
\begin{equation}
\label{eq:app_T_definition}
T_i^\star(\pi)(y\mid x)
=
\frac{\pi(y\mid x)\exp(f_i^\star(x,y))}
{Z_{i,\pi}(x)},
\qquad
Z_{i,\pi}(x)
:=
\sum_{u\in\Y}\pi(u\mid x)\exp(f_i^\star(x,u)).
\end{equation}
Then, under Assumption \ref{ass:exact_optimization}, $T_i^\star(\pi)$ is an exact population optimizer of
$\mathcal L_i(\cdot;\pi)$. Moreover, for every prompt $x$ and every pair of
responses $y,y'\in\Y$,
\begin{equation}
\label{eq:app_reference_increment}
\log\frac{T_i^\star(\pi)(y\mid x)}{\pi(y\mid x)}
-
\log\frac{T_i^\star(\pi)(y'\mid x)}{\pi(y'\mid x)}
=
f_i^\star(x,y)-f_i^\star(x,y').
\end{equation}
Therefore, the induced change in relative log-probability depends on the
source $i$ but not on the choice of reference policy $\pi$.
\end{lemma}

\begin{proof}
For convenience, write the population objective in score coordinates as
\begin{equation*}
\label{eq:app_R_i}
\mathcal R_i(f)
:=
\E_{(x,y^+,y^-)\sim P_i}
\left[
g\!\left(
\beta\left[f(x,y^+)-f(x,y^-)\right]
\right)
\right].
\end{equation*}
By Assumption~\ref{ass:exact_optimization}, $f_i^\star$ attains its minimum.

Because $\Y$ is finite and $\pi$ has full support,
$Z_{i,\pi}(x)$ in~\eqref{eq:app_T_definition} is finite and strictly
positive. Hence $T_i^\star(\pi)$ is a well-defined full-support policy. Its
log-policy ratio with respect to $\pi$ is
\begin{align}
h_{T_i^\star(\pi),\pi}(x,y)
&=
\log
\frac{T_i^\star(\pi)(y\mid x)}{\pi(y\mid x)}
\nonumber\\
&=
f_i^\star(x,y)-\log Z_{i,\pi}(x).
\label{eq:app_log_ratio_T}
\end{align}
Thus the reference policy affects the realized score through the
prompt-dependent normalization term $\log Z_{i,\pi}(x)$.

Since this term is constant across responses to the same prompt, it cancels
inside every preference comparison:
\begin{align*}
&
h_{T_i^\star(\pi),\pi}(x,y^+)
-
h_{T_i^\star(\pi),\pi}(x,y^-)
\\
&\qquad =
f_i^\star(x,y^+)-f_i^\star(x,y^-).
\end{align*}
Substituting into the DAA objective therefore gives
\begin{equation}
\mathcal L_i\!\left(T_i^\star(\pi);\pi\right)
=
\mathcal R_i(f_i^\star).
\label{eq:app_opt_value}
\end{equation}

For any other full-support policy $\tilde{\pi}$, its log-policy ratio
$h_{\tilde{\pi},\pi}$ defines a valid score function, and therefore
\begin{equation*}
L_i(\tilde{\pi};\pi)
=
R_i(h_{\tilde{\pi},\pi})
\geq
R_i(f_i^\star).
\end{equation*}
Together with~\eqref{eq:app_opt_value}, this proves that
$T_i^\star(\pi)$ is an exact population optimizer of
$\mathcal L_i(\cdot;\pi)$.

We now establish the reference-independence statement. For any two responses
$y,y'\in\Y$, subtracting~\eqref{eq:app_log_ratio_T} at $y'$ from the same
identity at $y$ gives
\begin{align}
&
\log
\frac{T_i^\star(\pi)(y\mid x)}{\pi(y\mid x)}
-
\log
\frac{T_i^\star(\pi)(y'\mid x)}{\pi(y'\mid x)}
\nonumber\\
&\qquad =
f_i^\star(x,y)-f_i^\star(x,y'),
\label{eq:app_reference_difference}
\end{align}
because $\log Z_{i,\pi}(x)$ cancels.

It remains to verify that the right-hand side is itself independent of the
reference policy. The objective $\mathcal R_i$ in~\eqref{eq:app_R_i} depends
on the source distribution $P_i$, $g$, and $\beta$, but not on $\pi$.
Moreover, Assumption~\ref{ass:exact_optimization} implies that any two
minimizers $f_i^\star$ and $\widetilde f_i^\star$ satisfy
\begin{equation*}
\widetilde f_i^\star(x,y)
=
f_i^\star(x,y)+a(x)
\end{equation*}
for some $a:\X\to\mathbb R$. Hence, for any $y,y'$,
\begin{align*}
\widetilde f_i^\star(x,y)
-
\widetilde f_i^\star(x,y')
&=
f_i^\star(x,y)+a(x)
-
f_i^\star(x,y')-a(x)
\\
&=
f_i^\star(x,y)-f_i^\star(x,y').
\end{align*}
Therefore the pairwise difference
\begin{equation*}
\Delta_i(x;y,y')
:=
f_i^\star(x,y)-f_i^\star(x,y')
\end{equation*}
is uniquely determined by source $i$ and does not depend on the choice of
minimizer.

Now let $\pi'$ be any other full-support reference policy. Repeating
\eqref{eq:app_reference_difference} with $\pi'$ gives
\begin{equation*}
\log
\frac{T_i^\star(\pi')(y\mid x)}{\pi'(y\mid x)}
-
\log
\frac{T_i^\star(\pi')(y'\mid x)}{\pi'(y'\mid x)}
=
\Delta_i(x;y,y').
\end{equation*}
The same quantity is therefore obtained for $\pi$ and $\pi'$. Since the two
reference policies were arbitrary, the change induced by source $i$ in the
relative log-probability of $y$ and $y'$ is independent of the reference
policy.

Finally, Assumption~\ref{ass:exact_optimization} also implies uniqueness of
the optimizer in policy space. Let $\bar\pi$ be any exact optimizer of
$\mathcal L_i(\cdot;\pi)$. Since
$T_i^\star(\pi)$ attains the minimum value
$\mathcal R_i(f_i^\star)$, the score $h_{\bar\pi,\pi}$ must also minimize
$\mathcal R_i$. By Assumption~\ref{ass:exact_optimization}, there therefore
exists $a:\X\to\mathbb R$ such that
\begin{equation*}
h_{\bar\pi,\pi}(x,y)
=
f_i^\star(x,y)+a(x).
\end{equation*}
Using the definition of the log-policy ratio,
\begin{equation*}
\bar\pi(y\mid x)
=
\pi(y\mid x)
\exp\!\left(f_i^\star(x,y)+a(x)\right).
\end{equation*}
Summing over $y\in\Y$ and using that $\bar\pi(\cdot\mid x)$ is normalized
gives
\begin{equation*}
1
=
\exp(a(x))
\sum_{y\in\Y}
\pi(y\mid x)\exp(f_i^\star(x,y))
=
\exp(a(x))Z_{i,\pi}(x).
\end{equation*}
Hence
\begin{equation*}
a(x)=-\log Z_{i,\pi}(x),
\end{equation*}
so that
\begin{equation*}
\bar\pi(y\mid x)
=
\frac{
\pi(y\mid x)\exp(f_i^\star(x,y))
}{
Z_{i,\pi}(x)
}
=
T_i^\star(\pi)(y\mid x).
\end{equation*}
Thus the optimal score is identified only up to a prompt-dependent additive
constant, while the corresponding optimal policy is unique.
\end{proof}

\subsection{Proof of Proposition~\ref{prop:composition}}

\begin{proposition}[Restatement of Proposition \ref{prop:composition}]
Let $S\subseteq\N$, let
$\sigma=(\sigma_1,\ldots,\sigma_m)$ be any ordering of its $m=|S|$ sources,
and consider the exact population sequential procedure
\begin{equation*}
\pi^{(0)}=\pi_0,
\qquad
\pi^{(k)}
\in
\argmin_{\pi}
\mathcal L_{\sigma_k}(\pi;\pi^{(k-1)}),
\qquad
k=1,\ldots,m.
\end{equation*}
Under Assumption~\ref{ass:exact_optimization}, its final policy is
\begin{equation}
\label{eq:app_composition_restatement}
\pi_S^\star(y\mid x)
=
\frac{
\pi_0(y\mid x)
\exp\!\left(
\sum_{i\in S}f_i^\star(x,y)
\right)
}{
\sum_{u\in\Y}
\pi_0(u\mid x)
\exp\!\left(
\sum_{i\in S}f_i^\star(x,u)
\right)
}.
\end{equation}
Consequently, the exact population policy depends only on the coalition $S$
and not on the ordering $\sigma$.
\end{proposition}

\begin{proof}
For $k=0,\ldots,m$, let
\begin{equation*}
S_k
:=
\{\sigma_1,\ldots,\sigma_k\},
\qquad
S_0:=\emptyset,
\end{equation*}
and define
\begin{equation}
\label{eq:app_Z_Sk}
Z_{S_k}(x)
:=
\sum_{u\in\Y}
\pi_0(u\mid x)
\exp\!\left(
\sum_{i\in S_k}f_i^\star(x,u)
\right).
\end{equation}
With the convention that an empty sum is zero,
$Z_{S_0}(x)=1$.

We prove by induction on $k$ that
\begin{equation}
\label{eq:app_induction_composition}
\pi^{(k)}(y\mid x)
=
\frac{
\pi_0(y\mid x)
\exp\!\left(
\sum_{i\in S_k}f_i^\star(x,y)
\right)
}{
Z_{S_k}(x)
}.
\end{equation}
For $k=0$, Equation~\eqref{eq:app_induction_composition} reduces to
$\pi^{(0)}=\pi_0$, so the claim holds.

Suppose now that Equation~\eqref{eq:app_induction_composition} holds at stage
$k-1$. The induction hypothesis implies that $\pi^{(k-1)}$ has full support,
so Lemma~\ref{lem:reference_invariance} applies with reference
$\pi^{(k-1)}$. Moreover, as established in its proof, the population optimizer
is unique in policy space. Therefore,
\begin{equation*}
\pi^{(k)}
=
T_{\sigma_k}^\star(\pi^{(k-1)}).
\end{equation*}
It follows that
\begin{align*}
\pi^{(k)}(y\mid x)
&=
\frac{
\pi^{(k-1)}(y\mid x)
\exp(f_{\sigma_k}^\star(x,y))
}{
\sum_{u\in\Y}
\pi^{(k-1)}(u\mid x)
\exp(f_{\sigma_k}^\star(x,u))
}
\\
&=
\frac{
\dfrac{\pi_0(y\mid x)}{Z_{S_{k-1}}(x)}
\exp\!\left(
\sum_{i\in S_k}f_i^\star(x,y)
\right)
}{
\dfrac{1}{Z_{S_{k-1}}(x)}
\sum_{u\in\Y}
\pi_0(u\mid x)
\exp\!\left(
\sum_{i\in S_k}f_i^\star(x,u)
\right)
}
\\
&=
\frac{
\pi_0(y\mid x)
\exp\!\left(
\sum_{i\in S_k}f_i^\star(x,y)
\right)
}{
Z_{S_k}(x)
}.
\end{align*}
This establishes the induction claim.

Taking $k=m$ gives Equation~\eqref{eq:app_composition_restatement}.
Since $S_m=S$, the final expression depends on the ordering $\sigma$ only
through the set of sources it contains. Because the sum
$\sum_{i\in S}f_i^\star(x,y)$ is invariant to permutations of $S$, every
ordering of the same coalition yields the same exact population policy.
\end{proof}

\subsection{Proof of Proposition~\ref{prop:reconstruction}}

\begin{proposition}[Restatement of Proposition \ref{prop:reconstruction}]
For each source $i$, let
$\pi_i^\star=T_i^\star(\pi_0)$ denote the exact singleton policy, and for any
coalition $S\subseteq\N$ define
\begin{equation}
\label{eq:app_reconstruction_restatement}
\widehat s_S^\star(x,y)
=
\log\pi_0(y\mid x)
+
\sum_{i\in S}
\left[
\log\pi_i^\star(y\mid x)
-
\log\pi_0(y\mid x)
\right].
\end{equation}
Under Assumption~\ref{ass:exact_optimization}, there exists a function
$c_S:\X\to\mathbb R$, independent of the response $y$, such that
\begin{equation}
\label{eq:app_reconstruction_target}
\widehat s_S^\star(x,y)
=
\log\pi_S^\star(y\mid x)+c_S(x).
\end{equation}
Consequently, for any two responses $y,y'\in\Y$ to the same prompt $x$,
\begin{equation}
\label{eq:app_pairwise_reconstruction}
\log\pi_S^\star(y\mid x)
-
\log\pi_S^\star(y'\mid x) =
\widehat s_S^\star(x,y)
-
\widehat s_S^\star(x,y').
\end{equation}
\end{proposition}

\begin{proof}
For each source $i$, define the singleton normalizer
\begin{equation*}
Z_i(x)
:=
\sum_{u\in\Y}
\pi_0(u\mid x)
\exp(f_i^\star(x,u)).
\end{equation*}
By Lemma~\ref{lem:reference_invariance},
\begin{equation*}
\pi_i^\star(y\mid x)
=
\frac{
\pi_0(y\mid x)\exp(f_i^\star(x,y))
}{
Z_i(x)
}.
\end{equation*}
Taking logarithms gives
\begin{equation}
\label{eq:app_singleton_logratio}
\log\pi_i^\star(y\mid x)
-
\log\pi_0(y\mid x)
=
f_i^\star(x,y)-\log Z_i(x).
\end{equation}
Substituting Equation~\eqref{eq:app_singleton_logratio} into
Equation~\eqref{eq:app_reconstruction_restatement} yields
\begin{equation}
\label{eq:app_hat_s_expanded}
\widehat s_S^\star(x,y)
=
\log\pi_0(y\mid x)
+
\sum_{i\in S}f_i^\star(x,y)
-
\sum_{i\in S}\log Z_i(x).
\end{equation}

On the other hand, Proposition~\ref{prop:composition} gives
\begin{equation}
\label{eq:app_piS_log}
\log\pi_S^\star(y\mid x)
=
\log\pi_0(y\mid x)
+
\sum_{i\in S}f_i^\star(x,y)
-
\log Z_S(x),
\end{equation}
where
\begin{equation*}
Z_S(x)
:=
\sum_{u\in\Y}
\pi_0(u\mid x)
\exp\!\left(
\sum_{i\in S}f_i^\star(x,u)
\right).
\end{equation*}
Comparing Equations~\eqref{eq:app_hat_s_expanded}
and~\eqref{eq:app_piS_log} gives
\begin{equation*}
\widehat s_S^\star(x,y)
=
\log\pi_S^\star(y\mid x)
+
\underbrace{
\left[
\log Z_S(x)
-
\sum_{i\in S}\log Z_i(x)
\right]
}_{=:c_S(x)}.
\end{equation*}
The term $c_S(x)$ depends on the coalition and the prompt, but not on the
response. This proves Equation~\eqref{eq:app_reconstruction_target}.

For $S=\emptyset$, $Z_S(x)=1$ and the sum over $i\in S$ is empty, so
$c_\emptyset(x)=0$ and the same identity holds.

Finally, subtracting Equation~\eqref{eq:app_reconstruction_target} for two
responses $y$ and $y'$ to the same prompt cancels $c_S(x)$, yielding
Equation~\eqref{eq:app_pairwise_reconstruction}.
\end{proof}

For completeness, the fixed-candidate normalization identity used immediately
after Proposition~\ref{prop:reconstruction} follows from the same result. For
any nonempty candidate set $\mathcal C(x)\subseteq\Y$, exponentiating
Equation~\eqref{eq:app_reconstruction_target} gives
\begin{equation*}
\exp\!\left(\widehat s_S^\star(x,y)\right)
=
\exp(c_S(x))\pi_S^\star(y\mid x).
\end{equation*}
The factor $\exp(c_S(x))$ is common to every response associated with prompt
$x$ and therefore cancels under normalization:
\begin{equation*}
\frac{
\exp\!\left(\widehat s_S^\star(x,y)\right)
}{
\sum_{u\in\mathcal C(x)}
\exp\!\left(\widehat s_S^\star(x,u)\right)
}
=
\frac{
\pi_S^\star(y\mid x)
}{
\sum_{u\in\mathcal C(x)}
\pi_S^\star(u\mid x)
}.
\end{equation*}
Thus the reconstruction preserves normalized distributions over a fixed
candidate set, as well as all pairwise score differences, rankings, and
argmax decisions within that set.

Likewise, if the evaluation protocol $\mathcal V$ satisfies the shift
invariance condition~\eqref{eq:shift_invariant_utility}, then
\begin{equation*}
\mathcal V(\widehat s_S^\star)
=
\mathcal V(\log\pi_S^\star+c_S)
=
\mathcal V(\log\pi_S^\star)
=
v(S).
\end{equation*}
This is the identity used in
Section~\ref{subsec:shapley_from_singletons} to recover every exact coalition
utility from the base policy and the $n$ singleton policies, without any
additional coalition-specific alignment.

\newpage

\section{Experimental details}
\label{sec:xp-details}
\label{app:experimental_details}

This appendix provides the complete experimental protocol used in
Sections~\ref{sec:finite-training-validation} and \ref{sec:shapley_application} and in Appendix \ref{sec:additional-experiments}.

\subsection{Preference data}
\label{app:data-details}

\paragraph{UltraFeedback sources.}
We use UltraFeedback \citep{cui2023ultrafeedback} and define one preference
source from each of the following nine prompt provenances:
\texttt{evol\_instruct},
\texttt{false\_qa},
\texttt{flan\_v2\_cot},
\texttt{flan\_v2\_flan2021},
\texttt{flan\_v2\_niv2},
\texttt{flan\_v2\_p3},
\texttt{sharegpt},
\texttt{truthful\_qa}, and
\texttt{ultrachat}.
These nine sources constitute the player set $\N$ throughout the experiments.

For each UltraFeedback prompt, we retain the highest-scoring valid completion
as the preferred response $y^+$ and the lowest-scoring valid completion as the
rejected response $y^-$. Examples with fewer than two valid scored
completions are discarded. Each resulting example is therefore a preference
triple $(x,y^+,y^-)$ associated with one source.

\paragraph{Training and final evaluation split.}
We first construct the complete normalized dataset containing the nine sources
and shuffle it once with a fixed seed. We then allocate $90\%$ of the examples to
the training pool and hold out the remaining $10\%$ for final evaluation.
The two sets are disjoint at the prompt-example level, and the final evaluation
pool is never used for training or checkpoint selection.

The $90\%$ training pool is partitioned by provenance to obtain one dataset
per source. For each source $i$, this source-specific pool is further split
into $75\%$ examples used for optimization and $25\%$ examples used for
checkpoint selection. We do not cap or rebalance the source training sets:
their original relative sizes are preserved.

\paragraph{Construction of the common evaluation dataset.}
The common evaluation dataset is constructed exclusively from the held-out
$10\%$ split. We first partition this split by source, retain at most $256$
preference triples from each source, concatenate the resulting source-specific
subsets, and shuffle the concatenated dataset with a fixed seed. We refer to this
dataset as \texttt{uf9\_eval}. No example in \texttt{uf9\_eval} is used for
training or optimization.

For the finite-training comparisons in
Section~\ref{sec:finite-training-validation}, we use the same fixed set $\mathcal E
=
\{(x_j,y_j^+,y_j^-)\}_{j=1}^{M},$ $M=512,$
drawn from \texttt{uf9\_eval}. The same $512$ preference triples are used for
all models, coalitions, source orders, reconstructions, and repeated training
runs.

\subsection{Models and optimization}
\label{app:model-training-details}

\paragraph{Base policies.}
We evaluate four instruction-tuned base policies:
\texttt{HuggingFaceTB/SmolLM-135M-Instruct},
\texttt{Qwen/Qwen2.5-0.5B-Instruct},
\texttt{meta-llama/Llama-3.2-1B-Instruct}, and
\texttt{microsoft/Phi-3.5-mini-instruct}.
We refer to them as SmolLM, Qwen2.5, Llama3.2, and Phi-3.5, respectively.

\paragraph{IPO training.}
The main finite-training experiments use IPO \citep{azar2024} with
$\beta=0.1$ for all four base policies. Each source is optimized for at most
$3$ epochs. We use a learning rate of $2\times10^{-5}$, per-device batch
size $4$, gradient accumulation over $4$ steps, bfloat16 precision, and a maximum sequence length of 1024. Except when the optimization seed is deliberately
varied to estimate retraining variability, all compared runs use the same
seed and identical optimization settings.

\paragraph{DPO training.}
The complementary DPO experiments use the sigmoid DPO loss and the same
optimization settings as above: a learning rate of $2\times10^{-5}$, per-device batch size $4$, gradient accumulation over 4 steps, a maximum sequence length of 1024,
bfloat16 precision, and an epoch ceiling of $3$.
The DPO temperature is $\beta=0.1$ for SmolLM and Qwen2.5. All DPO stages use the same stage-wise checkpoint-selection procedure described below.

\subsection{Checkpoint selection}
\label{app:checkpoint-selection}

Checkpoint selection is performed separately for every source optimization.
This applies both to singleton training and to \emph{every stage} of a
sequential alignment run.

Suppose a stage optimizes source $i$ from reference policy $\rho$. Candidate
checkpoints are evaluated only on the $25\%$ validation subset of source $i$.
For checkpoint selection, completion log-probabilities are length-normalized:
for a response $y=(y_1,\ldots,y_T)$,
\begin{align*}
\bar{\ell}_{\pi}(y\mid x)
=
\frac{1}{T}
\sum_{t=1}^{T}
\log \pi(y_t\mid x,y_{<t}).
\end{align*}
We then define the update margin relative to the current reference as
\begin{align*}
m_{\pi,\rho}(x,y^+,y^-)
=
\left[
\bar{\ell}_{\pi}(y^+\mid x)
-
\bar{\ell}_{\rho}(y^+\mid x)
\right]
-
\left[
\bar{\ell}_{\pi}(y^-\mid x)
-
\bar{\ell}_{\rho}(y^-\mid x)
\right].
\end{align*}

The primary checkpoint-selection criterion is the fraction of validation
examples for which $m_{\pi,\rho}>0$, i.e., the fraction for which the update
moves the preferred response upward relative to the rejected response.
Ties are broken first by the improvement in preference accuracy relative to
the reference policy and then by the mean update margin.

In a sequential run, once a checkpoint has been selected at stage $k$, this
checkpoint becomes both the initialization and the reference policy for stage
$k+1$. Hence checkpoint selection is performed anew after every source in the
sequence, using the held-out validation examples of the source currently being
optimized.

Length normalization is used only for checkpoint selection. All
sequence-level quantities used in the final finite-training evaluation and in
the coalition reconstruction use ordinary, non-length-normalized
complete-response log-probabilities.

\subsection{Singleton and sequential training}
\label{app:alignment-procedures}

\paragraph{Singleton models.}
For every source $i\in\N$, we train a singleton policy $\pi_i$ using only the
training portion of source $i$. Both the initialization and the reference
policy are the corresponding base policy $\pi_0$. The retained singleton is
the checkpoint selected according to the procedure above.

\paragraph{Sequential coalition models.}
For a coalition $S$ and an ordering
$\sigma=(\sigma_1,\ldots,\sigma_{|S|})$, we start from
$\pi^{(0)}=\pi_0$. At stage $k$, source $\sigma_k$ is optimized with
$\pi^{(k-1)}$ used both as initialization and reference, i.e., the selected checkpoint from the previous stage is used both to initialize the trainable policy and as the fixed reference policy in the DAA objective:
\begin{align*}
\pi^{(k)}
\leftarrow
\operatorname{DAA}\!\left(
D_{\sigma_k}^{\mathrm{train}};
\operatorname{init}=\pi^{(k-1)},
\operatorname{ref}=\pi^{(k-1)}
\right).
\end{align*}
Here, $\pi^{(k)}$ denotes the checkpoint selected on the validation subset of
source $\sigma_k$, rather than necessarily the final checkpoint of the
three-epoch training budget. The next stage starts from this selected
checkpoint.

\subsection{Finite-training validation}
\label{app:finite-training-details}

\paragraph{Evaluated coalitions and source orders.}
For each base model, we evaluate coalitions of sizes
$|S|\in\{3,6,9\}$. For every coalition, three prespecified source orderings
are trained under otherwise identical conditions. All three pairwise
comparisons between these orderings are included in the order-robustness
evaluation.

\paragraph{Coalitions, source orders, and repetitions.}
For the finite-training validation, we consider three nested coalitions
$S_3,S_6,S_9\subseteq\N$ of respective sizes $3$, $6$, and $9$:
\[
\begin{aligned}
S_3
&=
\{
\texttt{evol\_instruct},
\texttt{sharegpt},
\texttt{ultrachat}
\},
\\[1mm]
S_6
&=
\{
\texttt{evol\_instruct},
\texttt{false\_qa},
\texttt{flan\_v2\_flan2021},
\\[-1mm]
&\qquad
\texttt{sharegpt},
\texttt{truthful\_qa},
\texttt{ultrachat}
\},
\\[1mm]
S_9
&=
\N.
\end{aligned}
\]

We fix three global permutations of the complete set of nine sources:
\[
\begin{aligned}
\sigma_1 ={}&
\bigl(
\texttt{evol\_instruct},
\texttt{false\_qa},
\texttt{flan\_v2\_cot},
\texttt{flan\_v2\_flan2021},
\texttt{flan\_v2\_niv2},
\\[-1mm]
&\qquad
\texttt{flan\_v2\_p3},
\texttt{sharegpt},
\texttt{truthful\_qa},
\texttt{ultrachat}
\bigr),
\\[2mm]
\sigma_2 ={}&
\bigl(
\texttt{ultrachat},
\texttt{truthful\_qa},
\texttt{sharegpt},
\texttt{flan\_v2\_p3},
\texttt{flan\_v2\_niv2},
\\[-1mm]
&\qquad
\texttt{flan\_v2\_flan2021},
\texttt{flan\_v2\_cot},
\texttt{false\_qa},
\texttt{evol\_instruct}
\bigr),
\\[2mm]
\sigma_3 ={}&
\bigl(
\texttt{sharegpt},
\texttt{flan\_v2\_flan2021},
\texttt{evol\_instruct},
\texttt{ultrachat},
\texttt{flan\_v2\_cot},
\\[-1mm]
&\qquad
\texttt{truthful\_qa},
\texttt{false\_qa},
\texttt{flan\_v2\_p3},
\texttt{flan\_v2\_niv2}
\bigr).
\end{aligned}
\]

For a coalition $S$, we denote by $\sigma_k|_S$ the restriction of
$\sigma_k$ to the sources contained in $S$, obtained by deleting all sources
outside $S$ while preserving the relative order of the remaining ones.
Accordingly, the three orders used for the size-$3$ coalition are
\[
\begin{aligned}
\sigma_1|_{S_3}
&=
(\texttt{evol\_instruct},
 \texttt{sharegpt},
 \texttt{ultrachat}),\\
\sigma_2|_{S_3}
&=
(\texttt{ultrachat},
 \texttt{sharegpt},
 \texttt{evol\_instruct}),\\
\sigma_3|_{S_3}
&=
(\texttt{sharegpt},
 \texttt{evol\_instruct},
 \texttt{ultrachat}),
\end{aligned}
\]
while for the size-$6$ coalition they are
\[
\begin{aligned}
\sigma_1|_{S_6}
={}&
(\texttt{evol\_instruct},
 \texttt{false\_qa},
 \texttt{flan\_v2\_flan2021},
\\[-1mm]
&\qquad
 \texttt{sharegpt},
 \texttt{truthful\_qa},
 \texttt{ultrachat}),
\\[1mm]
\sigma_2|_{S_6}
={}&
(\texttt{ultrachat},
 \texttt{truthful\_qa},
 \texttt{sharegpt},
\\[-1mm]
&\qquad
 \texttt{flan\_v2\_flan2021},
 \texttt{false\_qa},
 \texttt{evol\_instruct}),
\\[1mm]
\sigma_3|_{S_6}
={}&
(\texttt{sharegpt},
 \texttt{flan\_v2\_flan2021},
 \texttt{evol\_instruct},
\\[-1mm]
&\qquad
 \texttt{ultrachat},
 \texttt{truthful\_qa},
 \texttt{false\_qa}).
\end{aligned}
\]
For $S_9=\N$, the three orders are simply
$\sigma_1,\sigma_2,\sigma_3$.

For each base model and each coalition
$S\in\{S_3,S_6,S_9\}$, we train one sequential coalition model under each of
the three orders $\sigma_1|_S$, $\sigma_2|_S$, and $\sigma_3|_S$.
Order sensitivity is evaluated using all three pairwise comparisons,
\[
(\sigma_1,\sigma_2),\qquad
(\sigma_1,\sigma_3),\qquad
(\sigma_2,\sigma_3),
\]
yielding three values of $d_{\mathrm{order}}$ for each model--coalition pair.
Reconstruction fidelity is evaluated separately against the explicitly
trained coalition produced under each of the three orders, yielding three
values of $d_{\mathrm{recon}}$ for the same model--coalition pair.

We designate $\sigma_{\mathrm{ref}}=\sigma_1$ as the global reference
ordering. For each model and coalition $S$, we repeat the complete sequential
training procedure following $\sigma_{\mathrm{ref}}|_S$ with $R=3$
independent optimization seeds. The coalition, source order, data splits,
hyperparameters, and checkpoint-selection procedure are held fixed across
these repetitions. The resulting three runs are used to estimate the
same-procedure retraining floor $d_{\mathrm{floor}}(S)$, which is shared by
all order and reconstruction comparisons associated with that
model--coalition pair.

\paragraph{Complete-response preference gaps.}
All finite-training discrepancies are evaluated on the common
$M=512$ examples in $\mathcal E$. For a policy $\pi$, we define
\begin{equation}
\label{eq:app_eval_gap}
G_{\pi}^{(j)}
=
\log\pi(y_j^+\mid x_j)
-
\log\pi(y_j^-\mid x_j).
\end{equation}
Here, $\log\pi(y\mid x)$ is the complete-response log-probability, obtained by
summing token log-probabilities over the response tokens. No length
normalization is applied.

For two vectors of preference gaps $G$ and $G'$, we use
\begin{equation}
\label{eq:app_gap_distance}
d(G,G')
=
\frac{1}{M}
\sum_{j=1}^{M}
\left|
G^{(j)}-G'^{(j)}
\right|.
\end{equation}

\paragraph{Order discrepancy.}
For two source orderings $\sigma$ and $\tau$ of the same coalition,
\begin{equation}
\label{eq:app_order_distance}
d_{\mathrm{order}}(S;\sigma,\tau)
=
d\!\left(
G_{\pi_{S,\sigma}},
G_{\pi_{S,\tau}}
\right).
\end{equation}

\paragraph{Reconstruction discrepancy.}
Given the trained singleton policies, the finite-training reconstruction is
\begin{equation}
\label{eq:app_empirical_reconstruction}
\widehat s_S(x,y)
=
\log\pi_0(y\mid x)
+
\sum_{i\in S}
\left[
\log\pi_i(y\mid x)
-
\log\pi_0(y\mid x)
\right].
\end{equation}
The corresponding reconstructed preference gap is
\begin{equation}
\widehat G_S^{(j)}
=
\widehat s_S(x_j,y_j^+)
-
\widehat s_S(x_j,y_j^-),
\end{equation}
and, for a sequential coalition trained under ordering $\sigma$,
\begin{equation}
\label{eq:app_recon_distance}
d_{\mathrm{recon}}(S;\sigma)
=
d\!\left(
\widehat G_S,
G_{\pi_{S,\sigma}}
\right).
\end{equation}

\paragraph{Retraining floor.}
For every model and coalition $S$, we fix one reference ordering
$\sigma_{\mathrm{ref}}$ and train this same ordered coalition independently
with $R=3$ optimization seeds, while keeping the data, order, and all other experimental
choices fixed. The retraining floor is
\begin{equation}
\label{eq:app_floor}
d_{\mathrm{floor}}(S)
=
\frac{2}{R(R-1)}
\sum_{1\le r<r'\le R}
d\!\left(
G_{\pi_{S,\sigma_{\mathrm{ref}}}^{(r)}},
G_{\pi_{S,\sigma_{\mathrm{ref}}}^{(r')}}
\right).
\end{equation}
Thus, with $R=3$, $d_{\mathrm{floor}}(S)$ is the mean of the three pairwise
distances between independent repetitions of the same ordered training
procedure.  The same value $d_{\mathrm{floor}}(S)$ is used to contextualize every order
and reconstruction discrepancy associated with the corresponding model and
coalition.

\subsection{Shapley valuation}
\label{app:shapley_details}

\paragraph{Candidate pools.}
We use the prompts $\mathcal X_{\mathcal E} =
\{x_j\}_{j=1}^{M}$
from the common held-out evaluation set. For each base policy $\pi_0$ and
each prompt $x$, we sample once a fixed pool of $K=16$
complete responses,
\begin{align*}
C(x)=\{y_1,\ldots,y_{16}\},
\qquad
y_k\sim\pi_0(\cdot\mid x).
\end{align*}
Generation uses fixed random seeds and a maximum of $64$ newly generated tokens per response. Consequently, for a given base model and prompt, the candidate pool is independent of the coalition being evaluated and is reused for all $2^9$ coalitions and all reward models.

We never sample from a reconstructed next-token coalition policy. For each coalition $S$, we instead score every fixed candidate using
\begin{equation}
\widehat s_S(x,y)
=
\log\pi_0(y\mid x)
+
\sum_{i\in S}
\left[
\log\pi_i(y\mid x)-\log\pi_0(y\mid x)
\right],
\end{equation}
and select
\begin{equation}
\label{eq:app_best_of_k}
\widehat y_S(x)
=
\argmax_{y\in C(x)}
\widehat s_S(x,y).
\end{equation}
The same candidate responses are therefore compared for every coalition.

\paragraph{Reward models.}
We evaluate each selected response using three external reward models: \texttt{Skywork/Skywork-Reward-V2-Qwen3-0.6B}, \texttt{Ray2333/GRM-Llama3.2-3B-rewardmodel-ft}, and \texttt{openbmb/UltraRM-13b}. We denote these models by $r_1$, $r_2$, and $r_3$, respectively.

Because the numerical scales of the reward models differ and to make comparisons across reward models visually interpretable, each reward is standardized over the same set of evaluated prompt-candidate pairs:
\begin{equation}
\widetilde r(x,y)
=
\frac{r(x,y)-\mu_r}{\sigma_r},
\end{equation}
where $\mu_r$ and $\sigma_r$ are the mean and standard deviation of $r(x,y)$ over all $x\in\mathcal X_{\mathcal E}$ and all $y\in C(x)$. These statistics are computed once for each reward model and kept fixed for all coalitions.

The utility of coalition $S$ is
\begin{equation}
\label{eq:app_shapley_utility}
\widehat v_r(S)
=
\frac{1}{|\mathcal X_{\mathcal E}|}
\sum_{x\in\mathcal X_{\mathcal E}}
\widetilde r\!\left(x,\widehat y_S(x)\right).
\end{equation}

\paragraph{Shapley computation.}
With $n=9$, we evaluate all
\begin{align*}
2^9=512
\end{align*}
coalitions. For each reward model $r$, the Shapley value of source $i$ is therefore computed exactly for the reconstructed game:
\begin{equation}
\phi_i(\widehat v_r)
=
\sum_{S\subseteq\N\setminus\{i\}}
\frac{|S|!(n-|S|-1)!}{n!}
\left[
\widehat v_r(S\cup\{i\})
-
\widehat v_r(S)
\right].
\end{equation}
No Monte Carlo approximation is used in the Shapley summation. The approximation in this paper instead replaces an explicitly trained coalition with its finite-training reconstruction from singleton policies.

Because utilities are computed from standardized rewards, $\phi_i(\widehat v_r)$ is expressed in standard-deviation units of
reward model $r$.
\newpage

\section{Additional experiments}
\label{sec:additional-experiments}

This appendix reports two complementary experiments. First, we repeat the finite-training validation in Section~\ref{sec:finite-training-validation} using DPO instead of IPO to verify that the empirical behavior supporting our reconstruction is not specific to the alignment loss used. Second, we compare sequential alignment with conventional pooled training on the same collections of sources. The latter comparison is not implied by our theory since sequential and pooled optimization correspond to different finite-training procedures. Rather, it serves as a sanity check that processing preference sources sequentially does not lead to substantially different alignment behavior from training jointly on their union.

\subsection{Complementary experiments with DPO}
\label{subsec:dpo-validation}

The theoretical results of Section~\ref{sec:efficient_shapley} apply to the class of Direct Alignment Algorithms described in Section~\ref{sec:preliminaries}, rather than specifically to IPO. Our main finite-training experiments instantiate this class with IPO. We therefore repeat the same validation with Direct Preference Optimization (DPO) \citep{rafailov2023} for two of the base models considered in the main experiments, \texttt{SmolLM-135M-Instruct} and
\texttt{Qwen2.5-0.5B-Instruct}.

We keep the experimental protocol of Section~\ref{sec:finite-training-validation} unchanged. All details are given in Appendix \ref{sec:xp-details}. The only substantive change is the direct-alignment loss, since DPO uses $g(t)=-\log\sigma(t)$ in place of the IPO objective.

\begin{figure}[h!]
    \centering
    \begin{subfigure}[t]{0.49\textwidth}
        \centering
        \includegraphics[width=\linewidth]{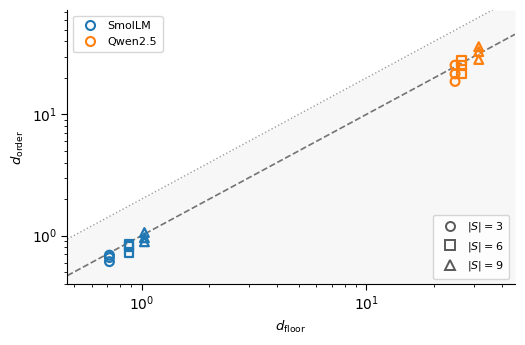}
        \caption{Order robustness}
        \label{fig:order_robustness_dpo}
    \end{subfigure}
    \hfill
    \begin{subfigure}[t]{0.49\textwidth}
        \centering
        \includegraphics[width=\linewidth]{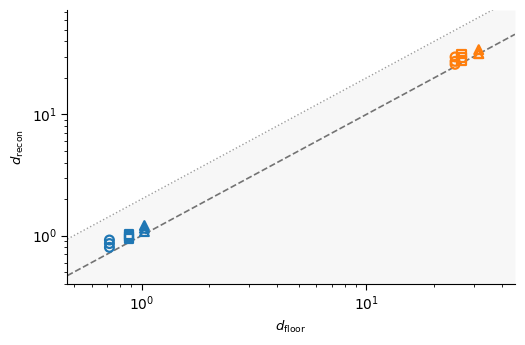}
        \caption{Reconstruction fidelity}
        \label{fig:reconstruction_fidelity_dpo}
    \end{subfigure}
    \caption{
    Finite-training validation with DPO for
    \texttt{SmolLM-135M-Instruct} and
    \texttt{Qwen2.5-0.5B-Instruct}.
    Left: source-order discrepancy $d_{\mathrm{order}}$. Right:
    reconstruction discrepancy $d_{\mathrm{recon}}$, both plotted against
    the corresponding retraining floor $d_{\mathrm{floor}}$.
    The dashed diagonal corresponds to equality with the retraining floor,
    while the dotted line marks twice that floor.
    }
    \label{fig:dpo-finite-training}
\end{figure}

Figure~\ref{fig:dpo-finite-training} reports the order and reconstruction discrepancies against the corresponding retraining floors. The plotting convention matches Figure~\ref{fig:finite_training_validation}.

The DPO results exhibit the same qualitative pattern as the IPO experiments in Section~\ref{sec:finite-training-validation}. Across both base models and the tested coalition sizes, source-order and reconstruction discrepancies remain of the same order as the corresponding retraining variability. Together with the IPO results, these experiments provide empirical evidence that the finite-training behavior underlying our reconstruction remains consistent with the broader class of DAAs covered by our theoretical analysis under Assumption \ref{ass:exact_optimization}.

\subsection{Sequential versus pooled preference optimization}
\label{subsec:sequential-vs-pooled}

We refer to \emph{pooled training} as the conventional procedure in which the
preference examples from all sources in a coalition $S$ are combined into a
single training set and optimized jointly in one run, with the base policy
$\pi_0$ used as the reference throughout training.

Our theory concerns sequential optimization and does not imply
that sequential and pooled training are identical under finite optimization. Even under exact
optimization, sequentially optimizing the source-specific objectives need not
recover the minimizer of their pooled objective. The experiment instead asks
whether, in the finite-training regime, replacing pooled training with the
sequential procedure underlying our reconstruction substantially changes the
resulting preference behavior.

We use \texttt{HuggingFaceTB/SmolLM-135M-Instruct} with IPO and consider the
same coalitions $S_3$, $S_6$, and $S_9$ as in the finite-training validation.
For each coalition, pooled training performs a single IPO optimization on the
union of its source datasets, using $\pi_0$ as the fixed reference. Sequential
training processes the same datasets source by source, using each updated
policy as the reference for the next stage. We evaluate the three sequential
source orders defined in Appendix~\ref{sec:xp-details}. Both procedures use
the same data and training hyperparameters, and each configuration is repeated
with $R=3$ independent optimization seeds.

We compare the resulting policies directly on the common held-out preference
set, without using an external reward model. For a policy $\pi$, let
\[
G_\pi(x,y^+,y^-)
=
\log\pi(y^+\mid x)-\log\pi(y^-\mid x),
\]
and recall the discrepancy
\[
d(\pi,\pi')
=
\frac{1}{M}\sum_{j=1}^M
\left|G_\pi^{(j)}-G_{\pi'}^{(j)}\right|.
\]
For each coalition, we report the largest sequential-pooled discrepancy over the three sequential orders. To make this quantity interpretable, we compare
it with the variability obtained by independently repeating the same training
procedure. Specifically, we define $d_{\mathrm{within}}$ as the largest mean
same-procedure discrepancy across the pooled and sequential configurations,
and report
\[
\rho_{\mathrm{seq/pool}}
=
\frac{d_{\mathrm{seq/pool}}}{d_{\mathrm{within}}}.
\]
Thus, $\rho_{\mathrm{seq/pool}} = 1$ means that the difference between
sequential and pooled training equals ordinary retraining
variability. 
We also report the fraction of held-out preference pairs on which the two
procedures induce opposite rankings.
\begin{table}[t]
\centering
\small
\begin{tabular}{ccccc}
\toprule
Coalition
&
$d_{\mathrm{seq/pool}}$
&
$d_{\mathrm{within}}$
&
$\rho_{\mathrm{seq/pool}}$
&
Ranking disagreement
\\
\midrule
$S_3$ & 0.73 & 0.60 & 1.20 & 1.3\% \\
$S_6$ & 1.01 & 0.78 & 1.28 & 2.2\% \\
$S_9$ & 1.30 & 0.90 & 1.43 & 3.0\% \\
\bottomrule
\end{tabular}
\caption{
Sequential versus pooled IPO training on
\texttt{SmolLM-135M-Instruct}.
$d_{\mathrm{seq/pool}}$ is the largest discrepancy over the three tested
sequential source orders, while $d_{\mathrm{within}}$ measures the largest
same-procedure retraining variability. Ranking disagreement is the fraction of
held-out preference pairs on which the two procedures prefer different
responses.}
\label{tab:sequential-pooled}
\end{table}

Table~\ref{tab:sequential-pooled} shows that the discrepancy between
sequential and pooled training remains close to the intrinsic variability of
retraining across all three coalition sizes. The ratio
$\rho_{\mathrm{seq/pool}}$ ranges from $1.20$ to $1.43$. The discrepancy increases moderately with coalition
size, but this increase is not accompanied by a substantial change in the
induced preference decisions. In fact, sequential and pooled policies disagree on only
$1.3\%$ to $3.0\%$ of held-out preference pairs. Thus, for
\texttt{SmolLM-135M-Instruct} in the regime considered here, sequential
training produces preference behavior that is close to pooled IPO relative to
the stochastic variability of training itself, while remaining measurably
distinct from it.
\newpage

\section{Additional figures}
\label{sec:additional-figures}
\begin{figure}[h!]
    \centering
    \vspace{-0.8em}
    \includegraphics[width=0.5\linewidth]{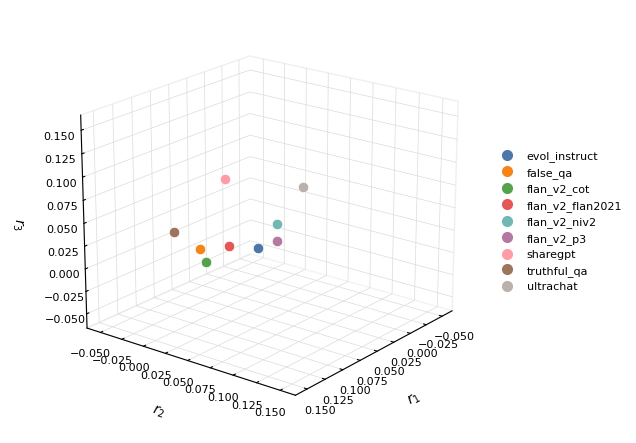}
    \caption{
    Shapley values of the $n=9$ preference sources for SmolLM under three standardized
    reward models. Each source $i$ is represented by
    $(\phi_{i}(\widehat{v}_{r_1}),
      \phi_{i}(\widehat{v}_{r_2}),
      \phi_{i}(\widehat{v}_{r_3})$.
      Along axis $r_k$, coordinates are expressed in standard-deviation units of the corresponding reward score.
    }
    \label{fig:shapley_3d_smollm}
    \vspace{-0.5em}
\end{figure}

\begin{figure}[h!]
    \centering
    \vspace{-0.8em}
    \includegraphics[width=0.5\linewidth]{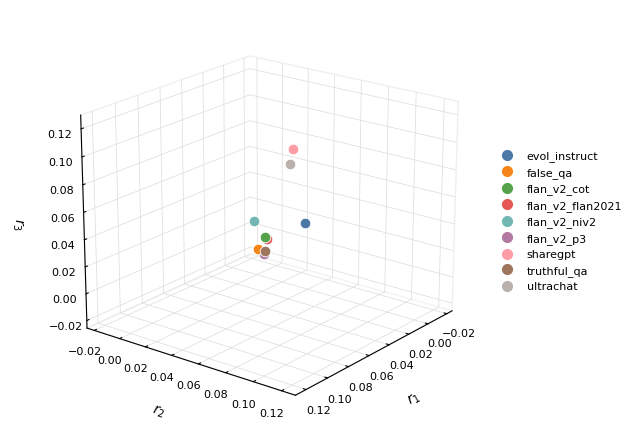}
    \caption{
    Shapley values of the $n=9$ preference sources for Llama3.2 under three standardized
    reward models. Each source $i$ is represented by
    $(\phi_{i}(\widehat{v}_{r_1}),
      \phi_{i}(\widehat{v}_{r_2}),
      \phi_{i}(\widehat{v}_{r_3})$.
      Along axis $r_k$, coordinates are expressed in standard-deviation units of the corresponding reward score.
    }
    \label{fig:shapley_3d_llama}
    \vspace{-0.5em}
\end{figure}

\begin{figure}[h!]
    \centering
    \vspace{-0.8em}
    \includegraphics[width=0.5\linewidth]{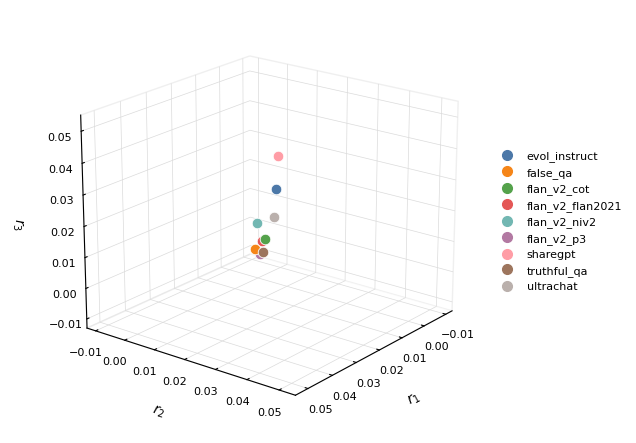}
    \caption{
    Shapley values of the $n=9$ preference sources for Phi-3.5 under three standardized
    reward models. Each source $i$ is represented by
    $(\phi_{i}(\widehat{v}_{r_1}),
      \phi_{i}(\widehat{v}_{r_2}),
      \phi_{i}(\widehat{v}_{r_3})$.
      Along axis $r_k$, coordinates are expressed in standard-deviation units of the corresponding reward score.
    }
    \label{fig:shapley_3d_phi}
    \vspace{-0.5em}
\end{figure}

This appendix complements Figure~\ref{fig:shapley_3d} by reporting the same Shapley-value representation for the other base models considered in our experiments. Figures~\ref{fig:shapley_3d_smollm}-\ref{fig:shapley_3d_phi} use exactly the evaluation protocol of Section~\ref{sec:shapley_application}.  Only the base model and corresponding singleton policies change.

\newpage

\section{Additional discussions}
\label{sec:additional-discussions}
\subsection{Extension to semivalue-based data valuation}
\label{subsec:semivalue_extension}

Our main analysis focuses on the Shapley value, but the reduction in alignment cost is not specific to the Shapley weighting. It applies directly to the broader family of semivalues introduced in Section~\ref{subsec:semivalue-based-data-valuation}.

Let $\omega=(\omega_0,\ldots,\omega_{n-1})$ be a probability distribution over coalition sizes, with $\omega_k\geq 0$ and $\sum_{k=0}^{n-1}\omega_k=1$. The corresponding semivalue of source $i$ can be written as
\begin{equation}
\label{eq:semivalue}
\psi_i^{\omega}(v)
=
\sum_{S\subseteq\N\setminus\{i\}}
\frac{\omega_{|S|}}{\binom{n-1}{|S|}}
\left[
v(S\cup\{i\})-v(S)
\right].
\end{equation}
Thus, $\omega_k$ determines the total weight assigned to marginal contributions where source $i$ is added to a coalition of size $k$, with this weight shared uniformly among coalitions of that size. The Shapley value is recovered by $\omega_k=1/n$ for every $k$. Other choices recover alternative semivalues. For example, the Banzhaf value \cite{banzhaf, wang2023data} corresponds to
\begin{equation*}
\omega_k=\frac{\binom{n-1}{k}}{2^{n-1}},
\end{equation*}
which assigns equal weight to every coalition
$S\subseteq\N\setminus\{i\}$.

The reconstruction developed in Section~\ref{sec:efficient_shapley} is independent of these weights. Under the exact population assumptions of that section, for every coalition $S$ the singleton reconstruction recovers the complete-response log-probabilities needed to evaluate $v(S)$, up to a prompt-dependent additive constant. Whenever the evaluation procedure is invariant to this constant, the
reconstructed utility satisfies
\begin{equation*}
\widehat v(S)=v(S)
\qquad
\text{for every } S\subseteq\N.
\end{equation*}
Substituting these reconstructed utilities into
Equation~\eqref{eq:semivalue} therefore gives
\begin{equation*}
\psi_i^{\omega}(\widehat v)
=
\psi_i^{\omega}(v)
\end{equation*}
for every source $i$ and every valid choice of semivalue weights $\omega$.

Consequently, changing from the Shapley value to another semivalue requires no additional alignment runs. The same base policy and the same $n$ singleton policies are sufficient. Only the weights used to aggregate coalition marginal contributions change. As for exact Shapley computation, evaluating an arbitrary semivalue exactly can still require utilities for exponentially many coalitions, but these utilities can be obtained through the same reconstruction rather than through coalition-specific alignment. Sampling-based estimators of semivalues can likewise be combined with our reconstruction when evaluating all coalitions is undesirable.

One distinction is important for interpretation. The efficiency identity used for the Shapley value,
\begin{equation*}
\sum_{i\in\N}\phi_i(v)=v(\N)-v(\emptyset),
\end{equation*}
does not hold for a general semivalue. Alternative semivalue weights therefore change not only the relative emphasis placed on different coalition sizes but also the axiomatic interpretation of the resulting data values. Our experiments focus on Shapley values because of this efficiency property. The choice among alternative semivalues is orthogonal to the coalition reconstruction developed in this work.

\subsection{From complete-response scores to next-token probabilities}
\label{subsec:next-token-gen-discussion}

Our reconstruction is defined at the level of complete responses. This section makes precise why this does not directly provide the next-token probabilities required for autoregressive generation.

Consider first the exact-optimization setting. By Proposition~\ref{prop:reconstruction}, for every coalition $S$,
\begin{align*}
\widehat s_S^\star(x,y)
=
\log \pi_S^\star(y\mid x)+c_S(x),
\end{align*}
where $c_S(x)$ does not depend on the response $y$. Hence $\widehat s_S^\star$ determines the complete-response distribution $\pi_S^\star(\cdot\mid x)$ up to its normalizing constant.

Let $u$ denote a response prefix, and let
\begin{align*}
\mathcal Y(u)
=
\{y\in\mathcal Y:\ y \text{ has prefix } u\}.
\end{align*}
For a token $a$ that can follow $u$, the next-token probability induced by the complete-response distribution is
\begin{equation}
\label{eq:next_token_from_sequence_scores}
\pi_S^\star(a\mid x,u)
=
\frac{
    \displaystyle
    \sum_{y\in\mathcal Y(ua)}
    \exp\!\left(\widehat s_S^\star(x,y)\right)
}{
    \displaystyle
    \sum_{b}
    \sum_{y\in\mathcal Y(ub)}
    \exp\!\left(\widehat s_S^\star(x,y)\right)
}.
\end{equation}
The prompt-dependent constant $c_S(x)$ and the global sequence-level normalizer cancel from this ratio. Nevertheless, Equation~\eqref{eq:next_token_from_sequence_scores} requires summing over all complete continuations of each possible next token. Thus, knowing how to score an individual complete response does not by itself provide an efficient way to compute the corresponding next-token distribution.

The distinction can also be seen directly from the autoregressive factorization of the reconstructed score. From Equation~\eqref{eq:practical_reconstruction},
\begin{align*}
\exp\!\left(\widehat s_S(x,y)\right)
=
\pi_0(y\mid x)
\prod_{i\in S}
\frac{\pi_i(y\mid x)}{\pi_0(y\mid x)}.
\end{align*}
Writing $y=(y_1,\ldots,y_T)$ and factorizing each model autoregressively gives
\begin{align*}
\exp\!\left(\widehat s_S(x,y)\right)
=
\prod_{t=1}^{T}
w_S(y_t\mid x,y_{<t}),
\end{align*}
where
\begin{equation}
\label{eq:local_combination_weight}
w_S(a\mid x,u)
=
\pi_0(a\mid x,u)
\prod_{i\in S}
\frac{\pi_i(a\mid x,u)}
     {\pi_0(a\mid x,u)}.
\end{equation}
It may therefore be tempting to obtain a next-token policy by normalizing $w_S(\cdot\mid x,u)$ locally. In general, however, this does not recover the conditional distribution associated with the complete-response scores.

To see why, define the continuation partition function
\begin{align*}
Z_S(x,u)
=
\sum_{y\in\mathcal Y(u)}
\prod_{t>|u|}
w_S(y_t\mid x,y_{<t}),
\end{align*}
with the natural convention that $Z_S(x,u)=1$ when $u$ is already a complete response. The next-token conditional corresponding to the globally normalized complete-response scores is then
\begin{equation}
\label{eq:next_token_with_continuation}
q_S(a\mid x,u)
=
\frac{
w_S(a\mid x,u)\,Z_S(x,ua)
}{
\displaystyle
\sum_b
w_S(b\mid x,u)\,Z_S(x,ub)
}.
\end{equation}
By contrast, locally combining and normalizing the singleton next-token probabilities would give
\begin{equation}
\label{eq:local_next_token_normalization}
\widetilde q_S(a\mid x,u)
=
\frac{
w_S(a\mid x,u)
}{
\displaystyle
\sum_b w_S(b\mid x,u)
}.
\end{equation}
Equations~\eqref{eq:next_token_with_continuation} and \eqref{eq:local_next_token_normalization} coincide only in special cases, for example, when the continuation partition $Z_S(x,ua)$ is the same for every possible next token $a$. There is no reason for this condition to hold for a general autoregressive language model since different next tokens lead to different prefixes and therefore to different distributions over all subsequent continuations.

This explains the scope of our reconstruction result. Evaluating $\widehat s_S(x,y)$ for a fixed complete response $y$ only requires teacher-forced evaluation of the base and singleton models along that response. Likewise, comparing or selecting among a fixed finite set of candidate
responses requires no sequence-level normalization. In contrast, sampling autoregressively from the complete-response distribution represented by $\widehat s_S$ would require computing, or approximating, the continuation partition functions in Equation~\eqref{eq:next_token_with_continuation} at every decoding step. For a general language model, these quantities involve summation over the tree of possible future continuations and cannot be obtained by simply normalizing the singleton probabilities at the current token.

For this reason, the present work uses the reconstruction only to score fixed complete responses and does not claim to reconstruct a generative coalition policy. Developing tractable approximations of the continuation marginalization required for direct generation is left for future work.

\end{document}